\documentclass{article}

\usepackage[preprint]{neurips_2025}

\usepackage{natbib}
\setcitestyle{numbers,square}
\usepackage[utf8]{inputenc} 
\usepackage[T1]{fontenc}    
\usepackage{hyperref}       
\usepackage{url}            
\usepackage{booktabs} 
\usepackage{multirow} 
\usepackage{siunitx}  
\usepackage{amsfonts}       
\usepackage{nicefrac}       
\usepackage{microtype}      
\usepackage[dvipsnames,table]{xcolor}
\usepackage{amsmath} 
\usepackage{amssymb}
\usepackage{graphicx}
\usepackage{wrapfig}
\usepackage{multirow}
\usepackage{array} 
\usepackage{geometry} 
\usepackage{tcolorbox}
\tcbuselibrary{skins, breakable}
\usepackage{listings}
\usepackage{textcomp}
\usepackage{fontawesome5} 
\usepackage{mathtools} 
\usepackage{enumitem}                     
\definecolor{deepPurple}{RGB}{102, 0, 153}

\NewDocumentCommand{\yi}
{ mO{} }{\textcolor{teal}{\textsuperscript{\textit{Yi}}\small[#1]}}

\title{\includegraphics[height=1.5em]{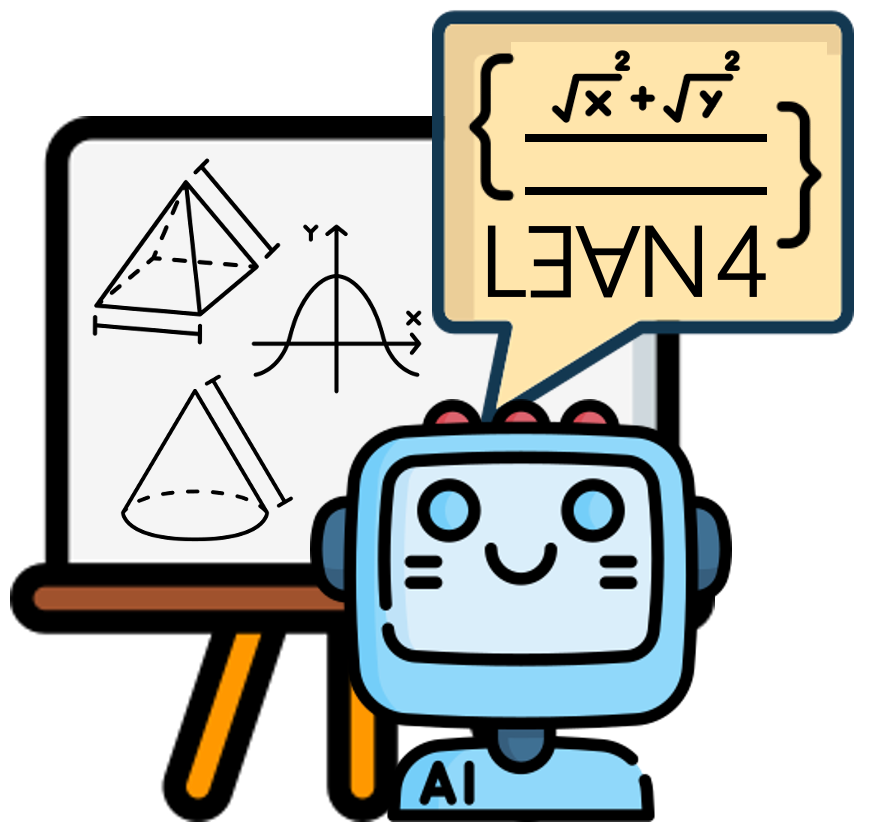} MATP-BENCH: Can MLLM Be a Good Automated Theorem Prover for Multimodal Problems?}

\author{%
  Zhitao He$^{\spadesuit}$ ~Zongwei Lyu$^{\spadesuit}$ ~Dazhong Chen$^{\clubsuit}$ ~Dadi Guo$^{\spadesuit}$ ~Yi R. (May) Fung$^{\spadesuit}$ \\
  $^{\spadesuit}$Hong Kong University of Science and Technology\\
  $^{\clubsuit}$Chinese University of Hong Kong (Shenzhen) \\
  \texttt{\{zhebu, yrfung\}@cse.ust.hk} \\
}

\begin{document}

\maketitle

\begin{abstract}
  Numerous theorems, such as those in geometry, are often presented in multimodal forms (e.g., diagrams). Humans benefit from visual reasoning in such settings, using diagrams to gain intuition and guide the proof process. Modern Multimodal Large Language Models (MLLMs) have demonstrated remarkable capabilities in solving a wide range of mathematical problems. However, the potential of MLLMs as Automated Theorem Provers (ATPs), specifically in the multimodal domain, remains underexplored. In this paper, we introduce the \textbf{M}ultimodal \textbf{A}utomated \textbf{T}heorem \textbf{P}roving benchmark (\textbf{MATP-BENCH}), a new Multimodal, Multi-level, and Multi-language benchmark designed to evaluate MLLMs in this role as multimodal automated theorem provers. MATP-BENCH consists of 1056 multimodal theorems drawn from high school, university, and competition-level mathematics. All these multimodal problems are accompanied by formalizations in Lean 4, Coq and Isabelle, thus making the benchmark compatible with a wide range of theorem-proving frameworks. MATP-BENCH requires models to integrate sophisticated visual understanding with mastery of a broad spectrum of mathematical knowledge and rigorous symbolic reasoning to generate formal proofs. We use MATP-BENCH to evaluate a variety of advanced multimodal language models. Existing methods can only solve a limited number of the MATP-BENCH problems, indicating that this benchmark poses an open challenge for research on automated theorem proving. The benchmark is publicly available at \url{https://matpbench.github.io}. 
\end{abstract}

\section{Introduction}
\label{intro}

\begin{figure}
    \centering
    \includegraphics[width=\textwidth]{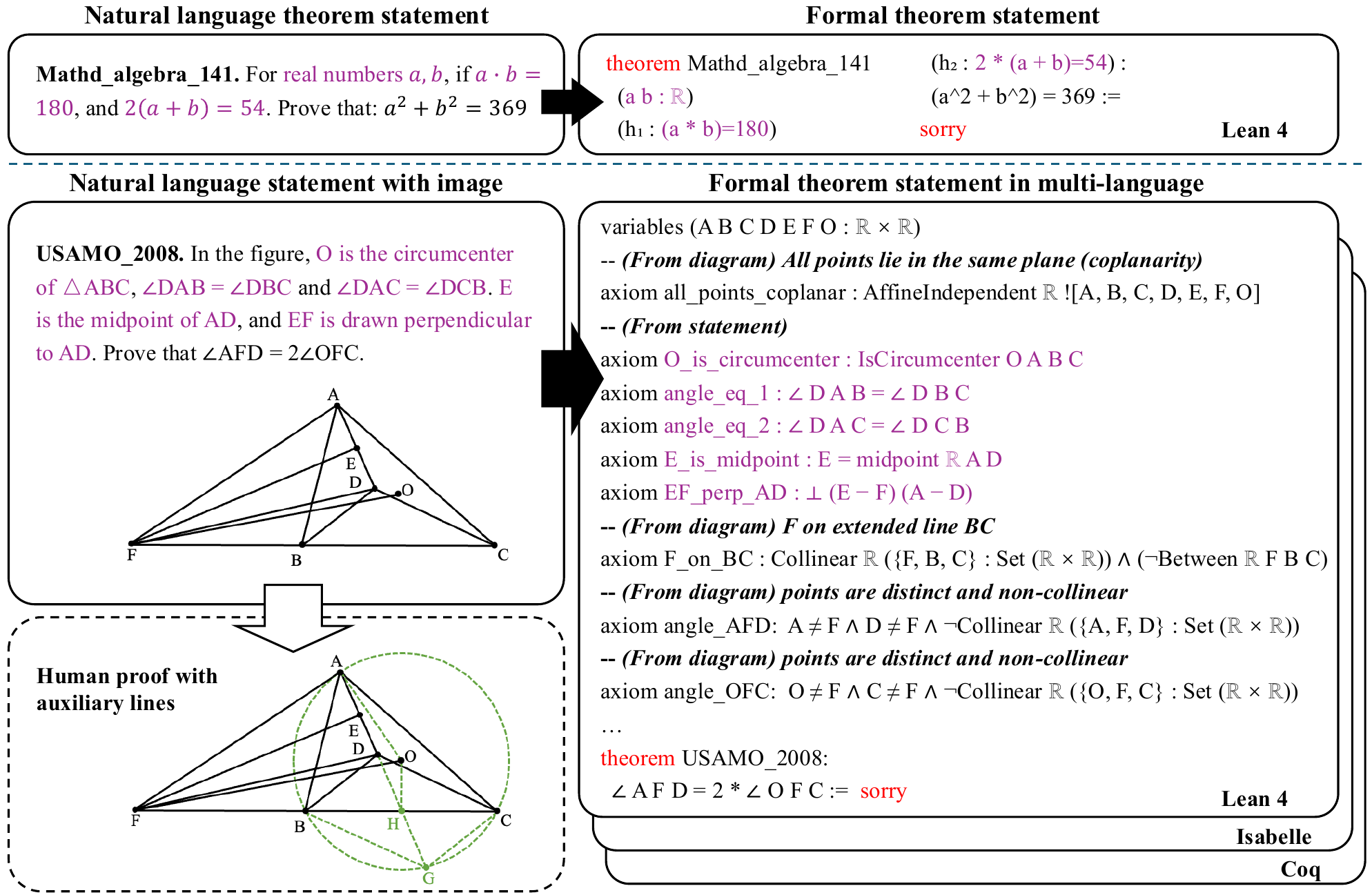}
    \caption{\small We illustrate the differences between traditional \textbf{ATP and MATP} through examples from miniF2F (above) and MATPBENCH (below). Multimodal theorems consist of an image paired with a natural language theorem statement, which complement each other to convey complete theorem information. Furthermore, additional auxiliary constructions are often essential for their proof (as shown in the bottom left subfigure). In traditional ATP, theorem formalization relies solely on textual statements (we use \textcolor{deepPurple}{purple} to indicate premises derived from the original statement), whereas MATP requires the model to extract critical premises not explicitly expressed in the text by analyzing accompanying diagrams (see \textbf{\textit{From diagram}} on the right). We provide formalized versions of all multimodal theorems in Lean4, Coq, and Isabelle.} 
    \label{fig:matp}
    \vspace{-0.3cm}
\end{figure}

In recent years, with the rapid advancement of large language models (LLMs), increasing attention has been devoted to their applications in automated theorem proving (ATP) \cite{azerbayev2023proofnet,lin2025goedel,tsoukalas2024putnambench,wang2024theoremllama,yang2023leandojo,xin2024deepseek,zheng2021minif2f}. LLMs is trained to generate rigorous formal proofs, which are then verified by systems such as Lean 4 \cite{moura2021lean}, Isabelle \cite{wenzel2008isabelle}, and Coq \cite{chlipala2013certified} to ensure the logical soundness of the reasoning process.

Although substantial progress has been made in automating the theorem proving, existing studies are generally limited to text-based inputs, and their potential for handling a wide range of theorems involving multimodal information has not yet been fully explored. Many mathematical problems require presentation and solution through multimodal information, such as geometric proofs, where non-textual elements like diagrams are crucial for problem understanding and reasoning, and mere textual descriptions are often insufficient. Current mainstream datasets, such as MiniF2F \cite{zheng2021minif2f} and ProofNet \cite{azerbayev2023proofnet}, mainly consist of textual theorems and lack support for multimodal theorem proving and multi-language proof verification ({\em e.g.,} in Coq, Isabelle). While LeanEuclid \cite{murphy2024autoformalizing} includes multimodal Euclidean geometry theorems, it primarily focuses on automatic formalization ({\em i.e.,} transforming human-written natural language proofs into formal proofs in Lean4) rather than on challenging the model's ability to autonomously generate formal proofs. AlphaGeometry \cite{seo2015solving}, on the other hand, can solve complex Olympiad geometry problems but is not multimodal and does not support verification in formal proof assistant systems like Lean 4.

Humans often rely on diagrams and visual structures to support mathematical reasoning. With the rapid progress of multimodal large language models (MLLMs), they have shown great potential in handling a wide range of mathematical problems \cite{lu2023mathvista,wang2024measuring,shi2024math,shao2024deepseekmath,wang2025letsreasonformallynaturalformal}, even those at Olympiad competition levels \cite{aimoprize, trinh2024solving, selsam2020imo}. However, the exploration of MLLMs as automated theorem provers for multimodal problems remains underexamined. Given their proficiency in multimodal comprehension, investigating their application in automated theorem proving could enable the handling of intricate proofs that rely on visual information, thereby pushing the boundaries of formal verification.

To this end, we present \textbf{M}ultimodal \textbf{A}utomated \textbf{T}heorem \textbf{P}roving benchmark (\textbf{MATP-BENCH}), a multimodal, multi-level, and multi-language benchmark for multimodal automated theorem proving (MATP). Our benchmark consists of 1056 multimodal theorems drawn from high school, university, and competition-level mathematics. Each theorem is accompanied by formal theorems in Lean 4 \cite{moura2021lean}, Isabelle \cite{wenzel2008isabelle}, and Coq \cite{chlipala2013certified}, making the benchmark compatible with a wide range of theorem-proving frameworks. 
As shown in Figure \ref{fig:matp}, each data sample in MATP-BENCH consists of an image, a natural language theorem statement, and formal theorem statements in three different languages. Compared to traditional ATP, MATP requires the integration of reasoning across both language and visual modalities. This is particularly crucial in geometry-related mathematical problems, which often rely on mathematical structures conveyed through images, such as topological relations, that are typically difficult to express precisely in natural language. For humans, the reasoning process in such tasks involves not only drawing intuitive insights from visual representations but also constructing auxiliary diagrams and identifying implicit structural relations. 


We use MATP-BENCH to evaluate a variety of multimodal language models. Our findings indicate that even current state-of-the-art models can only solve a limited number of problems, particularly when generating proofs in the Lean 4 language, where they perform poorly even for problems of only high school difficulty. Detailed analysis further reveals that: (1) All evaluated models commonly exhibit issues with incomplete understanding of problem information and the generation of invalid formal proof steps, indicating deficiencies in their visual-symbolic joint reasoning capabilities; (2) While models demonstrate some ability in converting natural language questions and geometric figures into formal statements, they still face significant challenges in subsequent complex logical reasoning and the construction of correct formal proofs; (3) Models can introduce steps or concepts involving auxiliary constructions (e.g., auxiliary lines) in proofs, but they fail to effectively utilize these constructions to substantively advance the proof process. 

\textbf{In summary, our key contributions are as follows}:

(1) We introduce the Multimodal Automated Theorem Proving benchmark (MATP-BENCH), which contains 1056 multimodal theorems drawn from high school, university, and competition-level mathematics. All these multimodal problems are accompanied by formalizations in Lean 4, Coq and Isabelle, thus making the benchmark compatible with a wide range of theorem-proving frameworks.

(2) We conduct extensive experiments on MATP-BENCH with six advanced multimodal language models of varying sizes. The experimental results show that even current state-of-the-art models can only solve a limited number of problems, particularly when generating proofs in the Lean 4 language, where they perform poorly even for problems of only high school difficulty.

(3) Our detailed analysis indicates that the primary bottleneck in current MATP task lies in symbolic reasoning and the construction of correct formal proofs. Although models have demonstrated an initial ability to introduce auxiliary constructions (e.g., auxiliary lines), they fail to effectively utilize these constructions to substantively advance the proof process. These findings offer valuable insights for future research. We release our
datasets and code publicly at \url{https://matpbench.github.io}. 
\section{Related Work}

\textbf{Automated Theorem Proving.} Automated theorem proving (ATP) has been a long-standing challenge in symbolic reasoning \cite{robinson2001handbook, bibel2013automated, zheng2021minif2f, liu2023fimo}, with substantial progress made in developing automated theorem provers \cite{polu2020generative,polu2022formal, lample2022hypertree,thakur2023context,azerbayev2023llemma,jiang2022draft,wang2024proving}. In recent years, multiple benchmarks \cite{jiang2021lisa, dwrensha_compfiles, ying2024lean, lin2025goedel} have been proposed for formal mathematical proof. MINIF2F \cite{zheng2021minif2f} features a diverse collection of 488 problems, each formalized in mainstream languages such as Lean 3 and Isabelle.  ProofNet \cite{azerbayev2023proofnet} comprises 371 parallel formal and natural language theorem statements with proofs, designed to evaluate autoformalization and theorem proving. PutnamBench \cite{tsoukalas2024putnambench} is a multi-language benchmark for theorem provers, featuring 1,692 formalizations of 640 Putnam Competition problems. All problems are manually formalized in Lean 4 and Isabelle, with many in Coq. FIMO \cite{liu2023fimo} contains 149 IMO Shortlisted Problems formalized in Lean, designed to advance automated theorem proving at the Olympiad level. LeanDojo \cite{yang2023leandojo} is a large-scale theorem-proving dataset sourced from mathlib, featuring a evaluation split that requires generalization to unseen premises. MATP-BENCH pushes the boundaries of formal verification by uniquely exploring richer multimodal automated theorem proving with diverse multi-language formalization proofs, unlike existing benchmarks that focus exclusively on purely text-based theorem proving in single formal languages.

\textbf{Multimodal Math Benchmarks.} Various benchmarks have been created to assess the mathematical reasoning capabilities of LLMs \cite{amini2019mathqa,cobbe2021training,mishra2022lila,frieder2023mathematical,hendrycks2020measuring,hendrycks2021measuring,ling2017program}, with a growing number of specialized evaluations for MLLMs \cite{lu2021inter,masry2022chartqa,lu2023mathvista,wang2024measuring}. 
GeoQA+ \cite{cao2022augmented}, UniGeo \cite{chen2022unigeo}, GEOS \cite{seo2015solving}, and Geometry3K \cite{lu2021inter} provide standardized benchmarks focused on plane geometry problem-solving. MMMU \cite{yue2024mmmu} offers a multi-disciplinary benchmark, incorporating a small subset of mathematical problems in multiple-choice format. MATHVERSE \cite{zhang2024mathverse} is a visual math benchmark with 2,612 core problems, generating 15K test samples across plane geometry, solid geometry, and functions. MATHVISTA \cite{lu2023mathvista} includes 6,141 examples to assess mathematical and visual reasoning across diverse tasks. MATH-Vision \cite{wang2024measuring} consists of 3,040 carefully selected problems with visual contexts, drawn from 19 real-world math competitions and spanning 12 grade levels. CMM-Math \cite{liu2024cmm} is a Chinese multimodal math dataset with over 28,000 high-quality samples across 12 grades, covering diverse problem types and providing detailed solutions. MV-MATH \cite{wang2025mv} is a multimodal benchmark featuring 2,009 multi-image questions (up to 8 images per question), categorized into three question types and 11 K-12 math subjects at three difficulty levels. Although significant progress has been made in multimodal mathematical benchmarking, the exploration of automated theorem proving in multimodal settings remains limited. To address this gap, we propose MATP-BENCH, a benchmark that evaluates the ability of MLLMs to integrate visual perception, mathematical reasoning, and symbolic manipulation to construct rigorous formal proofs. The detailed comparison between related benchmarks and MATPBENCH is shown in Table \ref{tab:compare}.

\section{Problem Formulation}
\label{sec3}

\begin{table}
\renewcommand{\arraystretch}{1.15} 
  \caption{Comparison of existing related benchmarks. MATP-BENCH is a \textbf{Multimodal, Multi-level, and Multi-language} benchmark designed to evaluate MLLMs as automated theorem provers.}
  \label{tab:compare}
  \centering
  \resizebox{\textwidth}{!}{ 
  \begin{tabular}{
    l c 
    >{\centering\arraybackslash}m{1.4cm} 
    >{\centering\arraybackslash}m{1.8cm} 
    >{\centering\arraybackslash}m{1.89cm} 
    >{\centering\arraybackslash}m{1.3cm} 
    >{\centering\arraybackslash}m{1.4cm} 
    c c c 
}
\toprule
\textbf{Benchmark} & \textbf{Size} & \textbf{Verifiable} & \textbf{Theorem Proving} & \textbf{Theorem formalization} & \textbf{Multi-modal} & \textbf{Multi-level} & \textbf{Lean} & \textbf{Isabelle} & \textbf{Coq} \\
\midrule
    \textbf{miniF2F} \cite{zheng2021minif2f} & 488 & \checkmark & \checkmark &  &     & \checkmark & \checkmark & \checkmark &   \\
    \textbf{ProofNet} \cite{azerbayev2023proofnet} & 371 & \checkmark & \checkmark &  &  &   & \checkmark &  & \\
    \textbf{Fimo} \cite{liu2023fimo}   & 149 & \checkmark & \checkmark &   &  & & \checkmark &  & \\
    \textbf{Geometry3K-test} \cite{lu2021inter}   & 601 & \checkmark &  &    & \checkmark & &  &  &\\
    \textbf{LeanEuclid} \cite{murphy2024autoformalizing}  & 173 & \checkmark &  &  \checkmark  & \checkmark & \checkmark & \checkmark &  &\\
    \textbf{PutnamBench} \cite{tsoukalas2024putnambench}  & 640 & \checkmark & \checkmark &  &  &   & \checkmark & \checkmark & \checkmark \\
    \textbf{AlphaGeometry-test} \cite{trinh2024solving}  & 30  & \checkmark & \checkmark &   &  \checkmark &   &   &   &   \\
    \textbf{ProverBench} \cite{ren2025deepseek}  &  325 & \checkmark & \checkmark &    &  & & \checkmark &  &\\
    \textbf{GeoTrust-test} \cite{fu2025trustgeogen}  & 240  & \checkmark &  &   & \checkmark  &  \checkmark &   &   &   \\
    \midrule
    \textbf{MATP-Bench}  & 1056 & \textcolor{blue}{\checkmark} & \textcolor{blue}{\checkmark} & \textcolor{blue}{\checkmark} & \textcolor{blue}{\checkmark} & \textcolor{blue}{\checkmark} & \textcolor{blue}{\checkmark} & \textcolor{blue}{\checkmark} & \textcolor{blue}{\checkmark} \\
    \bottomrule
  \end{tabular}}
\end{table}

\paragraph{Automated Theorem Proving (ATP).}
In the ATP task \cite{zheng2021minif2f, azerbayev2023proofnet, tsoukalas2024putnambench, liu2023fimo}, the system takes a \textbf{formalized theorem statement} \( T \) as input, as shown in the upper part of Figure \ref{fig:matp}. The goal is to generate a formal proof \( P \) such that:
\[
\texttt{Prover}_{ATP}(T) \rightarrow P,
\quad \text{where } \texttt{Check}(T, P) = \texttt{True}.
\]
Here, \texttt{Check} denotes the built-in proof verifier of the formal system (such as Lean, Coq, or Isabelle), which ensures that the generated proof \( P \) is a valid derivation of theorem \( T \).

\paragraph{Multimodal Automated Theorem Proving (MATP).}
In the MATP task, the input to the MATP system is a pair \( (I, S) \), where:
\begin{itemize}
\item \( I \) is a \textbf{multimodal input} (e.g., geometric figure);
\item \( S \) is a \textbf{natural language statement of the theorem} (not formalized).
\end{itemize}

As shown in the bottom part of Figure \ref{fig:matp}, the natural language statement \( S \) and the information from the multimodal input \( I \) are complementary, together forming a complete theorem. Hence, the model must first generate the complete formal theorem \( T \). Then, generates a valid formal proof \( P \). The entire MATP task can be summarized as:
\[
\texttt{Prover}_{MATP}(I, S) \rightarrow (T, P),
\quad \text{where } \texttt{Check}(T, P) = \texttt{True}.
\]

\paragraph{Preventing Modality Leakage.}
To avoid modality leakage, where the model could ignore visual inputs, we provide only natural language \( S \) and image \( I \), without the formalized theorem \( T \). This encourages the model to interpret and reason over multimodal inputs, as humans do when proving multimodal theorems. Furthermore, we incorporate a \textbf{formal theorem verification} task in the experimental setup to ensure that the formal theorems automatically generated by the multimodal model are consistent with the problem statement, rather than fabricating simple theorems arbitrarily.

\section{MATP-BENCH}
\label{bench}

MATP-BENCH is a new Multimodal, Multi-level, and Multi-language benchmark designed to evaluate MLLMs as automated theorem provers. MATP-BENCH consists of 1056 multimodal theorems drawn from high school, university, and competition-level mathematics. Each theorem is accompanied by formalization in Lean 4, Isabelle, and Coq, making the benchmark compatible with a wide range of theorem-proving frameworks.

\textbf{Multimodal Context and Multi-language Theorem.} Compared to previous automated theorem proving datasets, which typically contain only plain-text theorems, MATP-BENCH introduces concrete multimodal contexts to jointly evaluate models on visual understanding, mathematical reasoning, and symbolic manipulation. As shown in the lower part of Figure \ref{fig:matp}, each theorem consists of an image and a corresponding natural language description, which complement each other to form a complete statement. MATP-BENCH provides formalizations of these multimodal theorems in Lean 4, Isabelle, and Coq. To the best of our knowledge, MATP-BENCH is the first multimodal automated theorem proving benchmark covering all three of these languages. MATP-BENCH presents the following challenges: (1) \textbf{Visual Understanding}: The model must accurately extract key information from theorem-related images, akin to human perception, to construct formal theorem statements; (2) \textbf{Mathematical Reasoning}: It requires rigorous mathematical reasoning to derive complete proofs based on the provided natural language descriptions and images; (3) \textbf{Neural-Symbol Proof Generation}: The model must be proficient in these formal languages and capable of strictly translating the mathematical reasoning process into verifiable formal proof.


\begin{wraptable}{r}{0.63\textwidth}
\small
\centering
\caption{Statistics summary of MATP-BENCH. Counts and percentages are provided for each category.} 
\vspace{0.2cm}
\label{tab:data_summy}
\begin{tabular}{@{}llS[table-format=3.0]S[table-format=2.1,table-space-text-post=\%]}
\toprule
& {\textbf{Category}} & {\textbf{Count}} & {\textbf{Percentage}} \\
\midrule
\multirow{3}{*}{Level} & High School & 472 & 44.7\% \\
& College & 468 & 44.3\% \\
& Competition & 116 & 11.0\% \\
\midrule
\multirow{3}{*}{Type} & Plane Geometry & 937 & 88.7\% \\
& 3D Geometry & 73 & 6.9\% \\
& Analytic Geometry & 46 & 4.4\% \\
\midrule
\multirow{8}{*}{Topic} & Segment Relationships & 355 & 33.6\% \\
& Angle Relationships & 282 & 26.7\% \\
& Area Relationships & 222 & 21.0\% \\
& Circles and Tangents & 86 & 8.1\% \\
& Parallel and Perpendicular Lines & 38 & 3.6\% \\
& Similarity and Proportionality & 25 & 2.4\% \\
& Cyclic Quads \& Common Points & 20 & 1.9\% \\
& Other & 28 & 2.7\% \\
\bottomrule
\end{tabular}
\end{wraptable}

\textbf{Hierarchy and Diversity.} To comprehensively evaluate the potential of multimodal large language models (MLLMs) as automated theorem provers, MATP-BENCH meticulously features both clear hierarchy and rich diversity. Existing widely used benchmarks such as ProofNet \cite{azerbayev2023proofnet} mainly focus on basic mathematical problems from early undergraduate courses, FIMO \cite{liu2023fimo} is limited to high school mathematics, while PutnamBench \cite{tsoukalas2024putnambench} targets advanced mathematical reasoning at the undergraduate level. In contrast, as shown in Table \ref{tab:data_summy}, The problems in MATP-BENCH span three distinct educational stages—high school, university, and competitions—systematically covering a wide range of difficulty levels from elementary to advanced. Specifically, the high school and university problems are collected from publicly available multimodal math problem datasets \cite{lu2023mathvista, wang2024measuring, wang2025mv, lu2021inter}, while the competition problems are sourced from public Mathematical Olympiad examinations. Furthermore, we manually annotate the formal statements of each problem in three formal languages. This design ensures a thorough examination of models’ reasoning capabilities across different levels of complexity and mathematics problems. Moreover, the multimodal theorems in MATP-BENCH are primarily centered around the domains of geometry, such as plane geometry, 3D geometry, and analytic geometry. These theorems not only require models to demonstrate a solid grasp of fundamental geometric knowledge but also emphasize cross-modal understanding, complex reasoning, spatial modeling, and multi-step logical deduction, aiming to systematically evaluate models’ overall performance and depth of reasoning in structured mathematical tasks.

\textbf{Task Formulation.} As we mentioned in Section \ref{sec3}, we aim to achieve end-to-end multimodal automated theorem proving (Task 1), where the input is a natural language theorem and an image, and the output is a formal theorem and its proof, i.e. \texttt{Prover}\textsubscript{MATP}$(I, S) \rightarrow (T, P)$. Furthermore, to prevent the model from generating formal theorems that do not align with original problems, we separately set up multimodal theorem formalization (Task 2) for verification, which aligns with LeanEuclid \cite{murphy2024autoformalizing}. Thus, we divide the task into two progressively challenging sub-tasks: 
\begin{itemize}[topsep=0pt, itemsep=0pt, partopsep=0pt, parsep=0pt, leftmargin=*]
\item \textbf{Task 1: Multimodal Automated Theorem Proving}: This task aims to achieve end-to-end multimodal automated theorem proving similar to human provers, by directly generating a formalized theorem $T$ and its proof $P$ from multimodal informal input, i.e. \texttt{Prover}\textsubscript{Task1}$(I, S) \rightarrow (T, P)$. For example, the input of models are a natural language statement (USAMO 2008) and an image (geometric diagram), as shown in Figure \ref{fig:matp}. The required output is a formal theorem statement and a formalized valid proof. This presents a significant challenge as the model must first accurately formalize the theorem from multimodal input, and then subsequently construct a valid proof.

\item \textbf{Task 2: Multimodal Theorem Formalization}: The prover receives the multimodal question, and is required to formalize it into a precise theorem $T$, formally denoted as \texttt{Prover}\textsubscript{Task1}$(I, S) \rightarrow T$. For example, the model takes a natural language statement (USAMO 2008) and its corresponding image as input. Unlike task 1, the required output for task 2 is only the formal theorem statement (as shown on the right side of Figure \ref{fig:matp}), without the proof process. This task evaluates the model's ability to correctly understand and formalize information from both textual and visual modalities.
\end{itemize}

\textbf{Formalization Effort and Challenges.} Our formalization team consists of two doctoral students and several undergraduate students with backgrounds in advanced mathematics, computer science, and prior experience with formal proof assistants. The problems cover various question formats that we manually formalized on a case-by-case basis. Specifically: (1) Multiple-choice questions: we formalize these by using the correct answer and incorporating key information extracted from the accompanying image, reformulating the problem as a concrete theorem; (2) Fill-in-the-blank questions: the correct answer is directly filled into the problem statement, and the theorem is formalized by integrating relevant information present in the image; (3) Open-ended solution questions: we transform interrogative formulations into declarative statements based on the provided answer and combine them with visual cues from the image to construct a complete formal theorem. On average, fully formalizing a high school problem, a university problem, and a competition problem takes approximately 25, 30, and 15 minutes respectively (in one language). Among these, the descriptions of high school and university problems are relatively concise, with rich information contained in the images requiring preprocessing, while competition problems are more detailed and thus easier to formalize. Each formalization is reviewed by at least one other team member. In contrast to prior formalization tasks based solely on textual theorems, a central challenge in our setting lies in the incompleteness of the original natural language descriptions. Many essential assumptions are conveyed exclusively through diagrams. As a result, the formalization process requires manual identification and extraction of visual information, such as geometric structures, to reconstruct a complete and rigorous formal statement.


\section{Experiments}

\subsection{Experimental settings}
\label{experiment_setting}

\textbf{Methods:} we conduct extensive experiments on a wide variety of advanced multimodal large language models with different sizes. Specifically, the reasoning models \cite{li2025perception,qu2025survey,chen2025towards} include \texttt{OpenAI-o1} (\textbf{o1} for short) \cite{openai2024o3o4}, \texttt{Claude-3.7-Sonnet-Thinking} (\textbf{Claude-3.7} for short) \cite{anthropic2024claudesonnet}, and \texttt{Gemini-2.0-Flash-Thinking} (\textbf{Gemini-2.0} for short) \cite{google2025gemini25flash}; while the non-reasoning models include \texttt{GPT-4.1} (\textbf{GPT-4.1} for short) \cite{openai2024gpt41}, \texttt{Qwen2.5-VL-Instruct-70B} (\textbf{Qwen2.5-VL} for short) \cite{qwen2.5-VL}, and \texttt{Llama3.2-Vision-Instruct-11B} (\textbf{Llama3.2-V} for short) \cite{liu2023improved}. 

\begin{table}
\renewcommand{\arraystretch}{1.135} 
  \caption{Experimental results of \textbf{Multimodal Automated Theorem Proving} (Task 1), which requires model to generate both formalized theorem and proof. We adopt \textbf{pass@10} as the evaluation metric. We further present the experimental results of \textit{pass@n} (n=1, n=5) in Table \ref{task1_pass5} and \ref{task1_pass1}.} 
  \label{task1_pass10}
  \centering
  \resizebox{\textwidth}{!}{
\begin{tabular}{l|cccccc|c}
    \toprule
    \multirow{2}{*}{\textbf{Task 1}} & \multicolumn{6}{c}{\textbf{Method}} \vline & \multirow{2}{*}{\textbf{Avg.}} \\
    \cline{2-7}
    & \textbf{OpenAI-o1} & \textbf{Claude-3.7} & \textbf{Gemini-2.0} & \textbf{GPT-4.1} & \textbf{Qwen2.5-VL} & \textbf{Llama3.2-V}    & \\
    \midrule
    \multicolumn{7}{c}{\textbf{Lean 4 (pass@10)}} \\
    \midrule
    High school & 7.63 & 7.20 & 8.47 & \textbf{9.32} & 2.12 & 3.58 & 6.39   \\
    University & \textbf{4.70} & 3.85 & 2.14 & 2.99 & 1.50 & 1.92  & 2.85   \\
    Competition & 1.72 & 1.72 & 0.86 & \textbf{3.45} & 0.00 & 0.00 &  1.29   \\
    \rowcolor{cyan!3.8}
    Overall & 5.68 & 5.11 & 4.82  & \textbf{5.87} & 1.61 & 2.46 &  4.26  \\
    \midrule
    
    \multicolumn{7}{c}{\textbf{Coq (pass@10)}} \\
    \midrule
    High school & \textbf{28.37} & 22.47 & 14.76 & 28.27 & 7.59 & 6.96  & 18.07  \\
    University & 11.75 & \textbf{12.39} & 4.27 & 6.62 & 5.34 & 4.91  & 7.55   \\
    Competition & 5.45 & 4.31 & 1.72 & \textbf{9.48} & 0.86 & 0.17  & 3.67   \\
    \rowcolor{cyan!3.8}
    Overall & \textbf{19.43} & 16.92 & 8.71 & 16.64 & 3.59 & 7.37 & 12.15   \\
    \midrule
    
    \multicolumn{7}{c}{\textbf{Isabelle (pass@10)}} \\
    \midrule
    High school & \textbf{10.17}& 	8.90& 	7.84& 	11.23& 	4.03& 	3.60& 	 7.63   \\
    University & \textbf{7.48}& 	5.34& 	3.63& 	4.49& 	2.78& 	3.21 & 4.49   \\
    Competition & 0.86& \textbf{1.72}& 	0.00& 	2.59& 	0.00& 	0.00 & 0.86   \\
    \rowcolor{cyan!3.8}
    Overall & \textbf{6.75} &	5.9&	4.11&	6.39&	2.27&	2.45 & 4.65   \\
    
    \bottomrule
\end{tabular}}
\end{table}

\begin{table}[h]
\renewcommand{\arraystretch}{1.135} 
\centering
\caption{Experimental results of \textbf{Multimodal Theorem Formalization} (Task 2), which only require model to generate formalized theorem. We use GPT-4o as the judge and adopt \textbf{pass@10} as the evaluation metric. We present the experimental results of \textit{pass@n} (n=1, n=5) in Table \ref{task2_pass5} and \ref{task2_pass1}.}
\label{task2_pass10}
\resizebox{\textwidth}{!}{
\begin{tabular}{l|cccccc|c}
    \toprule
    \multirow{2}{*}{\textbf{Task 2}} & \multicolumn{6}{c}{\textbf{Method}} \vline & \multirow{2}{*}{\textbf{Avg.}} \\
    \cline{2-7}
    & \textbf{OpenAI-o1} & \textbf{Claude-3.7} & \textbf{Gemini-2.0} & \textbf{GPT-4.1} & \textbf{Qwen2.5-VL} & \textbf{Llama3.2-V}    & \\
    \midrule
    \multicolumn{7}{c}{\textbf{Lean 4 (pass@10)}} \\
    \midrule
    High school & 53.12 & \textbf{55.07}&44.02&47.58 & 26.84& 15.50  & 40.36  \\
    University & \textbf{61.28} & 60.42&56.11&58.19 & 33.33& 21.81 & 48.52  \\
    Competition & \textbf{63.32} & 61.64&59.40&56.03 & 38.66& 28.58  & 51.27  \\
    \rowcolor{cyan!3.8}
    Overall & \textbf{58.24} & 57.26&51.05&53.20& 31.46& 19.72  & 45.16 \\
    
    \midrule
    \multicolumn{7}{c}{\textbf{Coq (pass@10)}} \\
    \midrule
    High school & \textbf{42.65} & 39.07& 27.15& 37.03 & 16.73& 11.79  & 29.07  \\
    University & 45.64 & \textbf{51.22}& 30.00& 43.89 & 20.14& 16.39  & 34.56 \\
    Competition & 63.37 & 65.20 & 54.17& \textbf{68.35} & 38.66& 31.38  & 53.50 \\
    \rowcolor{cyan!3.8}
    Overall & 41.31 & \textbf{49.65} & 31.76& 43.13 & 20.64& 15.97  & 33.74 \\
    
    \midrule
    \multicolumn{7}{c}{\textbf{Isabelle (pass@10)}} \\
    \midrule
    High school & \textbf{51.88} & 49.78&35.79&42.10 & 21.94&17.42  & 36.49 \\
    University & 63.50 & 62.22&54.86&\textbf{65.73} & 31.81&23.89  & 50.34 \\
    Competition & \textbf{63.08} & 58.28&42.59&44.27 & 25.63&19.61  & 42.58 \\
    \rowcolor{cyan!3.8}
    Overall & \textbf{60.14} & 56.21&44.97&52.56 & 26.66&20.52  & 43.51 \\
    \bottomrule
\end{tabular}}
\end{table}

\textbf{Metrics:} For \textbf{Task 1}, which requires the model to generate both a correct formal theorem and its proof, we follow prior studies \cite{zheng2021minif2f, tsoukalas2024putnambench, liu2023fimo} and adopt \textbf{pass@$n$} (n=10) as the evaluation metric. This metric evaluates whether the prover can successfully complete a valid proof within $n$ attempts in the formal proof environment. For \textbf{Task 2}, which requires the model to formalize the multimodal theorem, we use GPT-4o as the judge to assess whether the formal theorem generated by the model is consistent with the our annotated ground truth. We also adopt \textbf{pass@$n$} as the evaluation metric. Details of the prompts for different tasks and evaluation can be found in Appendix \ref{prompt}.

\subsection{Main results}
\label{results}

\textbf{Lean 4.} End-to-end complete multimodal theorem formalization and proof generation (Task 1) is an extremely challenging task. The overall average performance across the models significantly declines with increasing theorem difficulty, dropping from 6.96\% at the high school level to 3.12\% at the university level and 2.08\% at the competition level, highlighting the models' shortcomings in handling complex geometric reasoning and generating rigorous formal proofs. Even for the currently strongest model, o1, its overall success rate (pass@10) is only 5.68\%. In contrast, Task 2 only required models to generate Lean 4 formal theorems based on multimodal input, and its overall success rate reached 46.81\%. The significantly higher success rate of Task 2 indicates that the models' ability to convert natural language descriptions and geometric figures into Lean 4 formal statements is relatively strong, suggesting that the primary bottleneck for Task 1 in Lean 4 is proof generation. 

\textbf{Coq.} In the Coq language tests, the overall success rate (pass@10) for the end-to-end complete multimodal theorem formalization and proof generation (Task 1) task is 12.86\%. Compared to the overall Task 1 success rate in Lean 4 (4.68\%), the models demonstrate stronger proof generation capability in Coq. We think that Coq's superior performance in Task 1 can be attributed to its more mature library of tactics, richer resources of formalized mathematics, and an imperative tactic language style that might be more suitable for current language models to learn. 
Nevertheless, the success rate still decreases with increasing theorem difficulty, although the drop appears less severe than in Lean 4. The o1 model ranks first with an overall success rate of 19.43\%, performing particularly well at the high school level (28.37\%). Task 2 (formalization only) in Coq has an overall success rate of 35.25\%, far higher than Coq's. Simultaneously, the models' overall performance in generating proofs within the Coq environment is superior to that in Lean 4.

\textbf{Isabelle.} The overall success rate of Task 1 (4.71\%) is comparable to that of Lean 4, but significantly lower than the overall success rate of Coq, indicating that models face similarly high challenges in performing end-to-end proof generation in the Isabelle environment as they do in Lean 4. We hypothesize this is primarily due to two factors: firstly, while the formulation style adopted for structured proofs in Isabelle is user-friendly for humans, its precise context management and high degree of structural requirements, making it difficult for models to produce complex proofs fully conforming to its syntax and logic. Secondly, although Isabelle incorporates powerful automation tools, model success likely depends not just on invoking these tools but also on generating appropriate intermediate steps or providing effective guidance.
The o1 model ranks first at the university difficulty level (7.48\%). The performance of the Qwen2.5 and Llama3.2 models in Task 1 under the Isabelle environment remains at the lower end. Models' ability to convert multimodal information into Isabelle formal statements is relatively strong, similar to Lean 4, and superior to Coq.

\begin{figure}[t]
    \centering
    \includegraphics[width=\textwidth]{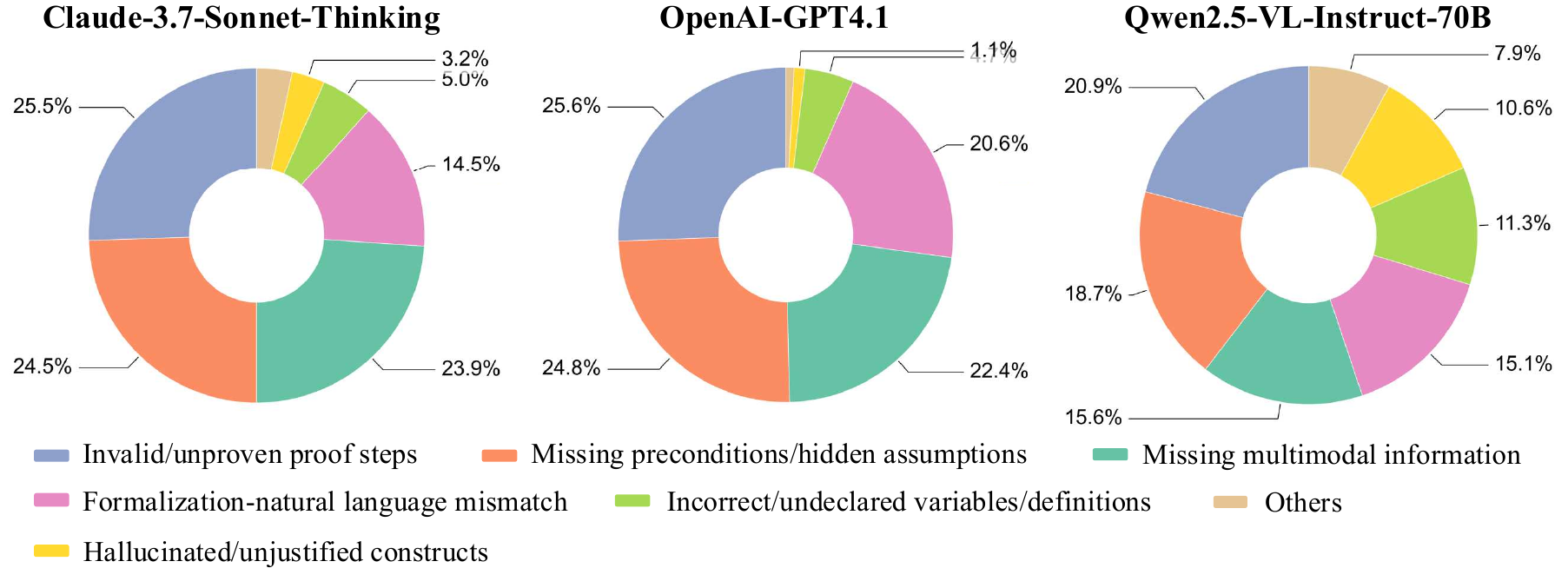}
    \caption{\small We perform an error analysis on the results of a reasoning model (Claude-3.7-Sonnet-Thinking) and two non-reasoning models (GPT4.1 and Qwen2.5-VL-Instruct-70B)
    , all three being competitive on MATP tasks (Lean 4), with the figure illustrating the seven most frequent error types.}
    \label{fig:error}
    \vspace{-0.3cm}
\end{figure}

\subsection{Analysis of Error Distribution in Multimodal Automated Theorem Proving}
We analyze the types of errors generated by multimodal models in theorem proving. As shown in Figure \ref{fig:error}, the results indicate that different models exhibit both common errors and specific issues. For the Claude model, the most significant errors are concentrated in invalid/unproven proof steps (24.9\%), missing preconditions/hidden assumptions (23.8\%), and missing unformalized information (underutilizing multimodal information, 23.3\%); these three categories collectively account for approximately 72\% of its total errors. Similarly, the error distribution of the GPT-4.1 model is very similar. This suggests that even for relatively better-performing models, the core challenges lie in complex logical reasoning and identifying and utilizing all implicit and explicit information required by the theorem. The error distribution of the Qwen2.5 model, however, differs. While missing unformalized information (20.9\%) and missing preconditions/hidden assumptions (18.7\%) are also major errors, the proportion of invalid/unproven proof steps is relatively lower (15.6\%), while more fundamental formalization errors such as missing/incorrect library imports (11.3\%) and incorrect/undeclared variables/definitions (10.6\%) are more prominent. This might indicate that while Qwen struggles with proof step errors, it also faces significant issues in generating basic code that conforms to the formal language specification. Overall, all models commonly exhibit problems with incomplete information understanding and broken reasoning chains.

\begin{figure}[ht]
    \centering
    \includegraphics[width=\textwidth]{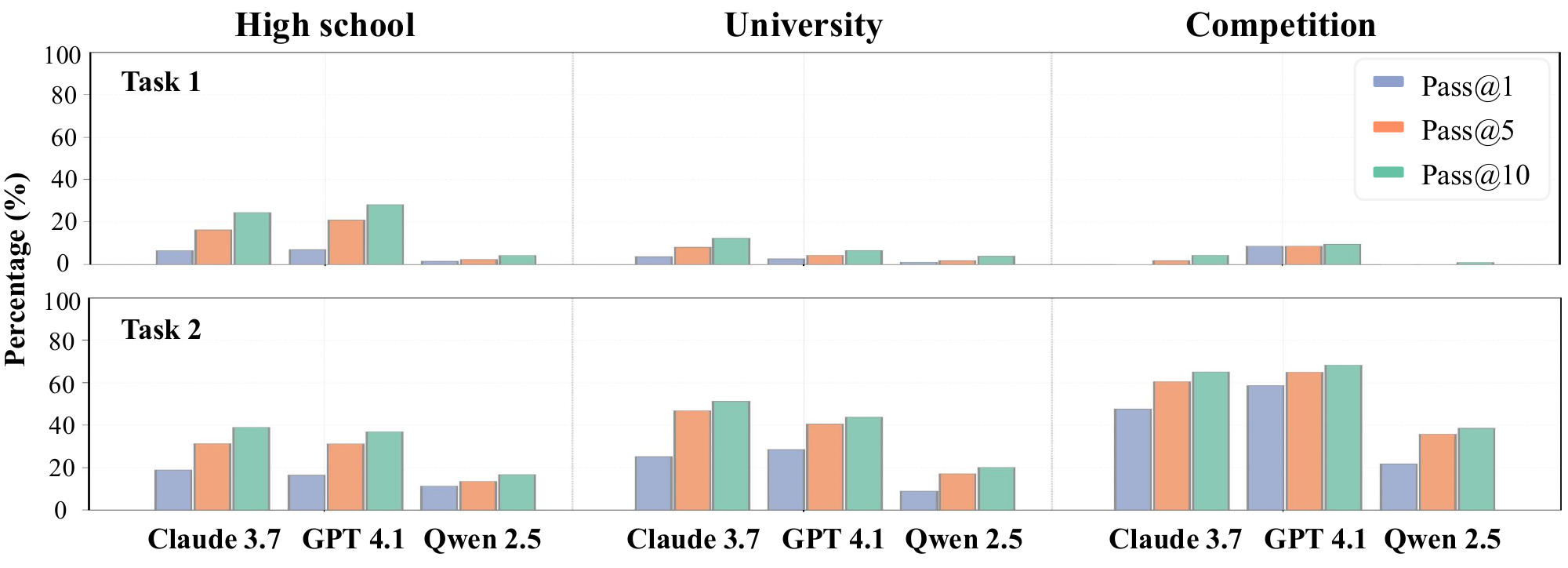}
    \caption{\small We present the performance of different MLLMs (Gemini-2.0-flash-thinking, OpenAI-GPT4.1, and Qwen2.5-VL-Instruct-70B) on multimodal theorem automated proving (Task 1) and theorem formalization (Task 2) across varying difficulty levels, evaluated using \textit{Pass@1, Pass@5, and Pass@10} metrics.}
    \label{fig:task12}
    \vspace{-0.3cm}
\end{figure}

\subsection{Main Bottleneck in Multimodal Automated Theorem Proving} 
As shown in Figure \ref{fig:task12}, the analysis of the pass@n performance of multimodal models in the Coq language shows that for both complete theorem formalization and proof generation (Task 1) and formalization only (Task 2), allowing models to generate more candidates (from pass@1 to pass@10) generally increases the success rate. However, the pass@n success rate for Task 1 is significantly lower than the pass@n success rate for Task 2 across all difficulty levels and models. For example, at the high school difficulty level, the highest pass@10 for Task 1 reaches 28.27\% for GPT-4.1, while the pass@10 for Task 2 for the same model reaches 37.03\%. Notably, the pass@n values for Task 2, especially at the competition difficulty level, can even remain at a relatively high level (68.35\% for GPT-4.1's pass@10). This may be attributed to the fact that the problem descriptions for competition-level theorems are often more complete and precise, providing clearer formalization basis for the models, even if the proof process heavily relies on complex steps like constructing auxiliary lines (which is precisely why the Task 1 success rate is extremely low at this difficulty). This large performance gap between Task 1 and Task 2 is consistent across all pass@n settings, demonstrating that models have shown a certain ability in converting natural language descriptions and geometric figures into Coq formal statements (Task 2 pass@n is relatively high), but still face significant challenges in the subsequent complex logical reasoning and constructing formal proofs. 

\begin{wrapfigure}{L}{0.5\textwidth} 
    \centering 
    \includegraphics[width=\linewidth]{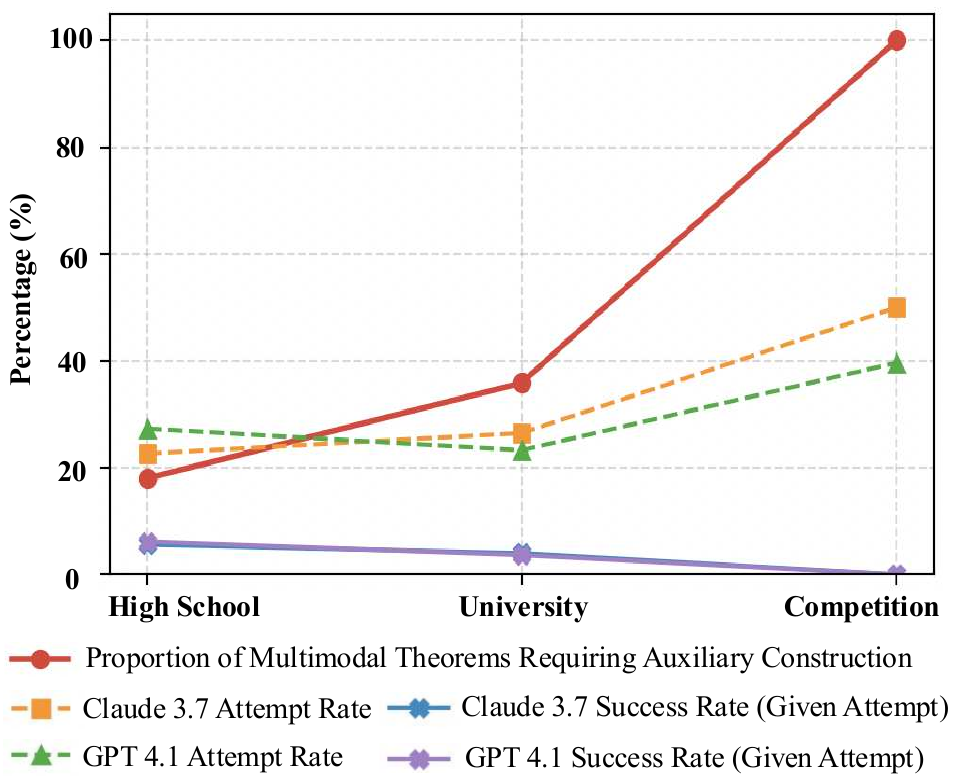} 
    \caption{\small Auxiliary construction analysis by question difficulty level and model, evaluated using \textit{Pass@10}.}
    \label{fig:construction}
\end{wrapfigure}

\subsection{Model Capability in Generating Auxiliary Constructions for Multimodal Theorems}

A characteristic distinguishing multimodal theorem proving from pure text theorem proving is that many theorems require the construction of auxiliary lines to aid thinking, especially problems at the competition level. Therefore, we further investigate the models' ability to construct auxiliary lines during the proof process. As shown in Figure \ref{fig:construction}, with the increase in theorem difficulty, the proportion of problems requiring auxiliary construction significantly rises, confirming the importance of auxiliary construction for solving high-difficulty geometric theorems. Claude 3.7 and GPT 4.1, which perform best in Task 1 and Task 2, attempt to perform auxiliary constructions when generating proofs, and this attempt rate also increases with difficulty, which indicates that the models possess a certain degree of autonomous auxiliary construction capability and awareness. However, the success rate of these proofs containing auxiliary constructions (Figure \ref{fig:construction} \textcolor{Cerulean}{blue} and \textcolor{Orchid}{purple} solid lines) is very low. 
This prominently indicates that while models can introduce steps or concepts involving auxiliary constructions in proofs, they cannot effectively utilize these constructions to advance the proof process. However, recent research shows that multimodal models can be augmented by visual prompts \cite{hu2024visual, shtedritski2023does, yang2023set}. For example, Visual Sketchpad \cite{hu2024visual} demonstrates this potential by providing MLLMs with a sketching interface and drawing tools, which offers a promising direction for multimodal theorem proving. 

\section{Conclusion}

In this paper, we introduce \textbf{M}ultimodal \textbf{A}utomated \textbf{T}heorem \textbf{P}roving benchmark (\textbf{MATP-BENCH}), a new multimodal, multi-level, and multi-language benchmark designed to evaluate Multimodal Large Language Models (MLLMs) as automated theorem provers. MATP-BENCH consists of 1056 multimodal theorems drawn from high school, university, and competition-level mathematics. All these multimodal problems are accompanied by formalizations in Lean 4, Coq and Isabelle. Our experimental results demonstrate significant variability in the performance of various mainstream MLLMs. Our findings highlight the current capabilities and limitations of state-of-the-art MLLMs in automated theorem proving, and indicate that the primary bottleneck in current MATP task lies in the construction of correct formal proofs, presenting clear directions for future research. 


\bibliographystyle{plainnat}
\bibliography{reference}

\newpage
\appendix

\section{Examples of questions at different levels}
\begin{figure}[ht]
    \centering
    \includegraphics[width=0.9\textwidth]{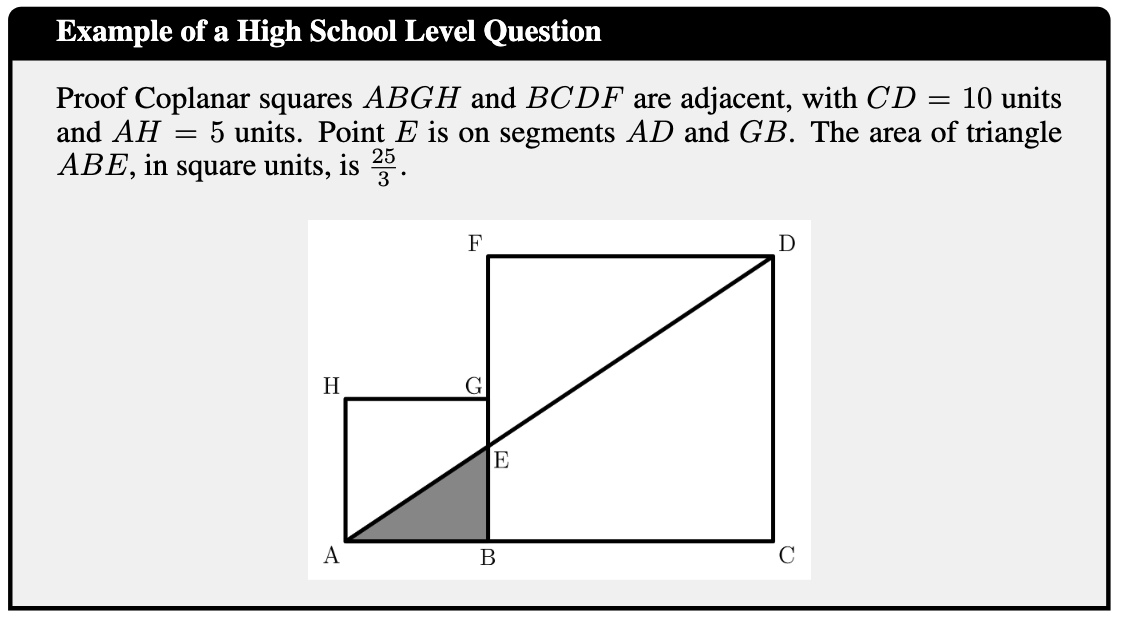}
    \caption{\small An example of a high school level mathematics problem, requiring the calculation of a triangle's area within a configuration of two adjacent squares (ABGH and BCDF) of differing side lengths.}
    \label{fig:high_example}
    \vspace{-0.3cm}
\end{figure}

\begin{figure}[ht]
    \centering
    \includegraphics[width=0.9\textwidth]{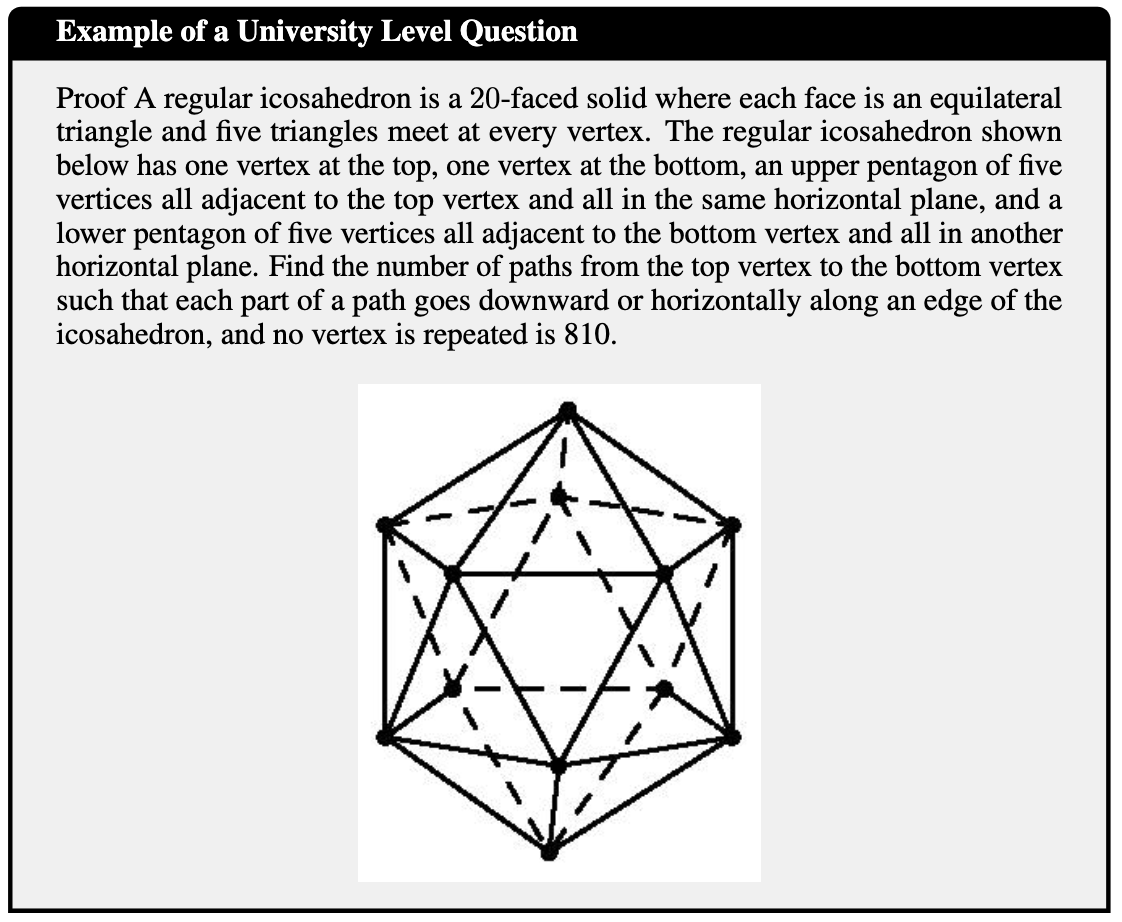}
    \caption{\small An example of a university level mathematics problem, requiring the determination that the number of non-repeating paths from the top to bottom vertex of a regular icosahedron, under downward or horizontal movement constraints, is 810.}
    \label{fig:university_example}
    \vspace{-0.3cm}
\end{figure}

In Figure \ref{fig:high_example} and Figure \ref{fig:university_example}, we show high school and university-level problems respectively, with Figure \ref{fig:matp} featuring competition-level questions.

\begin{table}[ht]
\renewcommand{\arraystretch}{1.15} 
  \caption{Experimental results of \textbf{Multimodal Automated Theorem Proving} (Task 1), which requires model to generate both formalized theorem and proof. We adopt \textbf{pass@5} as the evaluation metric.}
  \label{task1_pass5}
  \centering
  \resizebox{\textwidth}{!}{
\begin{tabular}{l|cccccc|c}
    \toprule
    \multirow{2}{*}{\textbf{Task 1}} & \multicolumn{6}{c}{\textbf{Method}} \vline & \multirow{2}{*}{\textbf{Avg.}} \\
    \cline{2-7}
    & \textbf{OpenAI-o1} & \textbf{Claude-3.7} & \textbf{Gemini-2.0} & \textbf{GPT-4.1} & \textbf{Qwen2.5-VL} & \textbf{Llama3.2-V}   & \\
    \midrule
    \multicolumn{7}{c}{\textbf{Lean 4 (pass@5)}} \\
    \midrule
    High school & 4.03 & 3.39 & \textbf{5.51} & 5.08 & 1.48 & 2.54   & 3.67 \\
    University & 2.78 & 1.71 & 1.50 & \textbf{3.38} & 0.86 & 1.72  & 1.99 \\
    Competition & 0.00 & 0.86 & 0.00 & \textbf{1.72} & 0.00 & 0.00   & 0.43  \\
    \rowcolor{cyan!3.8}
    Overall & 3.03 & 2.27 & 3.12 & \textbf{3.69} & 1.04 & 1.61   & 2.46  \\
    
    \midrule
    \multicolumn{7}{c}{\textbf{Coq (pass@5)}} \\
    \midrule
    High school & 18.78 & 16.24 & 8.65 & 20.89 & 3.58 & 4.43   & 12.09 \\
    University & \textbf{6.84} & 8.12  & 3.85 & 4.27 & 2.56 & 3.21  & 4.81  \\
    Competition & 1.72 & 1.72 & 0.86 & \textbf{8.62} & 0.00 & 0.00  & 2.15  \\
    \rowcolor{cyan!3.8}
    Overall & 11.63 & 11.08 & 5.67 & 12.19 & 2.65 & 3.40   & 7.78  \\
    
    \midrule
    \multicolumn{7}{c}{\textbf{Isabelle (pass@5)}} \\
    \midrule
    High school & 6.78 &	5.30	 &4.45 &	7.84 &	2.54 &	1.91   & 4.80  \\
    University & \textbf{4.70} &	3.42 &	2.14 &	3.82 &	1.92 &	2.35  & 3.06  \\
    Competition & 0.86 &	1.72 &	0.00	 &\textbf{2.59} &	0.00 &	0.00   & 0.86  \\
    \rowcolor{cyan!3.8}
    Overall & 4.11&	3.48&	2.2	&\textbf{4.97}&	1.49&	1.42  & 2.94\\
    
    \bottomrule
\end{tabular}}
\end{table}

\begin{table}[ht]
\renewcommand{\arraystretch}{1.15} 
  \caption{Experimental results of \textbf{Multimodal Automated Theorem Proving} (Task 1), which requires model to generate both formalized theorem and proof. We adopt \textbf{pass@1} as the evaluation metric.}
  \label{task1_pass1}
  \centering
  \resizebox{\textwidth}{!}{
\begin{tabular}{l|cccccc|c}
    \toprule
    \multirow{2}{*}{\textbf{Task 1}} & \multicolumn{6}{c}{\textbf{Method}} \vline & \multirow{2}{*}{\textbf{Avg.}} \\
    \cline{2-7}
    & \textbf{OpenAI-o1} & \textbf{Claude-3.7} & \textbf{Gemini-2.0} & \textbf{GPT-4.1} & \textbf{Qwen2.5-VL} & \textbf{Llama3.2-V}   & \\
    \midrule
    \multicolumn{7}{c}{\textbf{Lean 4 (pass@1)}} \\
    \midrule
    High school & 2.75 & 2.54 & 3.18 & \textbf{3.39} & 0.85 & 1.48   & 2.36   \\
    University & 1.50 & 0.85 & 0.43 & \textbf{2.85} & 0.43 & 0.64  & 1.12  \\
    Competition & 0.00 & \textbf{0.86} & 0.00 & 0.00 & 0.00 & 0.00   & 0.14  \\
    \rowcolor{cyan!3.8}
    Overall & 1.89 & 1.52 & 1.61 & \textbf{2.56} & 0.57 & 0.95   & 1.52  \\
    
    \midrule
    \multicolumn{7}{c}{\textbf{Coq (pass@1)}} \\
    \midrule
    High school & \textbf{10.13} & 6.54 & 3.59 & 6.96 & 1.48 & 3.16   & 5.31   \\
    University & \textbf{4.91} & 3.63 & 1.28 & 2.56 & 1.07 & 1.44   & 2.48 \\
    Competition &  0.00 & 0.00 & 0.86 & \textbf{7.62} & 0.00 & 0.00   & 1.41  \\
    \rowcolor{cyan!3.8}
    Overall & \textbf{6.72} & 4.54 & 2.27 & 5.20 & 1.13 & 2.08  & 3.66 \\
    
    \midrule
    \multicolumn{7}{c}{\textbf{Isabelle (pass@1)}} \\
    \midrule
    High school & \textbf{4.24}& 	3.60& 	2.97& 	4.87& 	1.27	& 0.85  & 2.97  \\
    University & \textbf{3.28}& 	2.14& 	0.85& 	2.14& 	1.07& 	1.28  & 1.79 \\
    Competition & 0.00 & \textbf{1.72}& 	0.00& 	0.86& 	0.00	& 0.00	  & 0.43  \\
    \rowcolor{cyan!3.8}
    Overall &\textbf{3.18}& 	1.91& 	1.27	& 2.62	& 0.78& 	0.71  & 1.75\\
    
    \bottomrule
\end{tabular}}
\end{table}

\section{Multimodal Automated Theorem Proving}\label{appendix:task1}
Tables \ref{task1_pass5} and \ref{task1_pass1} present the experimental results for Task 1, evaluating multimodal automated theorem proving using pass@5 and pass@1 metrics, respectively, across Lean 4, Coq, and Isabelle formal languages and three difficulty levels. Comparing the two tables, it is evident that providing models with more attempts (pass@5 vs pass@1) generally leads to higher success rates across all models, formal languages, and difficulty levels, highlighting the benefit of multiple decoding attempts in this task. Analyzing the pass@5 results in Table 5, among the individual models, GPT-4.1 consistently ranks among the top performers under pass@5, showing notable strength in handling higher difficulty levels. Gemini-2.0, Qwen2.5-VL, and Llama3.2 generally achieve lower pass rates across most tasks and difficulty levels under pass@5.

The pass@1 results in Table 6, which assesses the model's ability to generate a correct proof on the very first attempt, are considerably lower across the board. The relative ranking of models shifts for some languages under this stricter metric. GPT-4.1 still demonstrates relative strength at the competition level even at pass@1 in Coq and Isabelle, suggesting some capability for direct high-difficulty solutions. The performance difference between pass@5 and pass@1 highlights that while all models benefit from retries, some models, appear to leverage multiple attempts more effectively to find a successful proof compared to their initial attempt performance, whereas others, like o1 in Coq and Isabelle, are relatively stronger at generating a correct proof on the first try.

\begin{table}[ht]
\renewcommand{\arraystretch}{1.15} 
  \caption{Experimental results of \textbf{Multimodal Theorem Formalization} (Task 2), which only requires model to generate formalized theorem. We adopt \textbf{pass@5} as the evaluation metric.}
  \label{task2_pass5}
  \centering
  \resizebox{\textwidth}{!}{
\begin{tabular}{l|cccccc|c}
    \toprule
    \multirow{2}{*}{\textbf{Task 2}} & \multicolumn{6}{c}{\textbf{Method}} \vline & \multirow{2}{*}{\textbf{Avg.}} \\
    \cline{2-7}
    & \textbf{OpenAI-o1} & \textbf{Claude-3.7} & \textbf{Gemini-2.0} & \textbf{GPT-4.1} & \textbf{Qwen2.5-VL} & \textbf{Llama3.2-V}   & \\
    \midrule
    \multicolumn{7}{c}{\textbf{Lean 4 (pass@5)}} \\
    \midrule
    High school & \textbf{49.64} & 47.58&39.36&42.92 & 24.70&13.58 & 36.29   \\
    University & \textbf{60.14} & 58.89&52.36&55.00 & 30.83&21.27  & 46.42   \\
    Competition & \textbf{60.52} & 58.84&55.47&52.11 & 36.42&25.90  & 48.21   \\
    \rowcolor{cyan!3.8}
    Overall & \textbf{55.48} & 53.82&46.88&49.27 & 28.38 &55.17 & 48.17  \\
    
    \midrule
    \multicolumn{7}{c}{\textbf{Coq (pass@5)}} \\
    \midrule
    High school & \textbf{38.95} & 31.40&23.45&31.27 & 13.54&9.87  & 24.75   \\
    University & 45.83 & \textbf{46.89}&25.83&40.69 & 17.22&14.31  & 31.79  \\
    Competition & \textbf{63.58} & 60.52&53.23&65.04 & 35.86&30.26  & 51.42   \\
    \rowcolor{cyan!3.8}
    Overall & \textbf{44.85} & 39.91&27.77&39.14 & 18.06&14.07  & 30.63 \\
    
    \midrule
    \multicolumn{7}{c}{\textbf{Isabelle (pass@5)}} \\
    \midrule
    High school & 43.16 & 42.37&30.44&\textbf{45.52} & 19.34 & 15.63   & 32.74   \\
    University & \textbf{61.67} & 60.83&48.61&57.36 & 29.03 & 21.67   & 46.53  \\
    Competition & \textbf{57.72} & 54.91&37.54&38.10 & 21.45 & 16.81   & 37.75   \\
    \rowcolor{cyan!3.8}
    Overall & \textbf{52.39} & 50.91&39.26&45.46 & 23.96 & 18.43   & 38.40   \\
    
    \bottomrule
\end{tabular}}
\end{table}

\begin{table}[ht]
\renewcommand{\arraystretch}{1.15} 
  \caption{Experimental results of \textbf{Multimodal Theorem Formalization} (Task 2), which only requires model to generate formalized theorem. We adopt \textbf{pass@1} as the evaluation metric.}
  \label{task2_pass1}
  \centering
  \resizebox{\textwidth}{!}{
\begin{tabular}{l|cccccc|c}
    \toprule
    \multirow{2}{*}{\textbf{Task 2}} & \multicolumn{6}{c}{\textbf{Method}} \vline & \multirow{2}{*}{\textbf{Avg.}} \\
    \cline{2-7}
    & \textbf{OpenAI-o1} & \textbf{Claude-3.7} & \textbf{Gemini-2.0} & \textbf{GPT-4.1} & \textbf{Qwen2.5-VL} & \textbf{Llama3.2-V}   & \\
    \midrule
    \multicolumn{7}{c}{\textbf{Lean 4 (pass@1)}} \\
    \midrule
    High school & \textbf{33.05} & 30.72 & 24.14 & 26.19 & 13.71 &7.68   & 24.80  \\
    University & 45.61 & \textbf{45.82} & 35.14 & 37.22 & 20.56 &13.8  & 35.07 \\
    Competition & \textbf{43.15} & 39.78 & 36.98 & 33.06 & 24.66 &19.05  & 35.16  \\
    \rowcolor{cyan!3.8}
    Overall & \textbf{40.24} & 38.21 & 30.41 & 31.82 & 17.94 &11.67   & 30.52  \\
    
    \midrule
    \multicolumn{7}{c}{\textbf{Coq (pass@1)}} \\
    \midrule
    High school & \textbf{27.66} & 18.92&18.10&16.46 & 11.38 & 8.68  & 17.73   \\
    University & \textbf{35.49} & 25.28&12.92&28.61 & 8.89 & 5.36   & 21.94   \\
    Competition & 57.26 & 47.63&34.18&\textbf{58.72} & 21.85 & 12.61   & 41.72   \\
    \rowcolor{cyan!3.8}
    Overall & \textbf{34.62} & 24.88&17.57&26.36 & 11.43 & 8.35   & 22.13 \\
    
    \midrule
    \multicolumn{7}{c}{\textbf{Isabelle (pass@1)}} \\
    \midrule
    High school & 23.68 & \textbf{26.31}&15.77&16.73 & 10.97 & 8.78   & 18.29   \\
    University & 51.19 & 49.86&27.36&\textbf{53.33} & 18.19 & 13.47   & 36.35  \\
    Competition & \textbf{35.66} & 31.70&21.85&26.34 & 12.89 & 8.95   & 25.46   \\
    \rowcolor{cyan!3.8}
    Overall & 37.18 & \textbf{39.76}&21.56&29.55 & 14.38 & 10.87  & 26.97   \\
    
    \bottomrule
\end{tabular}}
\end{table}

\section{Multimodal Theorem Formalization}
\label{appendix:task2}
Tables \ref{task2_pass5} and \ref{task2_pass1} present the experimental results for Multimodal Theorem Formalization (Task 2), which evaluates models on generating formalized theorems using pass@5 and pass@1 metrics across Lean 4, Coq, and Isabelle. As expected, pass@5 scores consistently exceed pass@1 scores, indicating that multiple attempts improve formalization accuracy. However, compared to full proof generation (Task 1), the relative increase from pass@1 to pass@5 appears less dramatic for Task 2, suggesting that models capable of formalizing a theorem often do so successfully on earlier attempts. Overall, for theorem formalization, models achieve considerably higher pass rates than for proof generation, highlighting that generating the correct theorem statement is a less challenging task than generating the complete proof.

Analyzing the results, o1 demonstrates particular strength in first-attempt formalization (pass@1), frequently leading in Coq and Isabelle. Models like Claude-3.7 and GPT-4.1 show competitive performance in specific languages or difficulty tiers, while others generally trail. These results indicate that while current models are significantly better at theorem formalization than full proof generation, their ability to accurately formalize theorems still varies depending on the specific formal language and problem complexity, with o1 showing notable capabilities in this task.

\begin{figure}[ht]
    \centering
    \includegraphics[width=\textwidth]{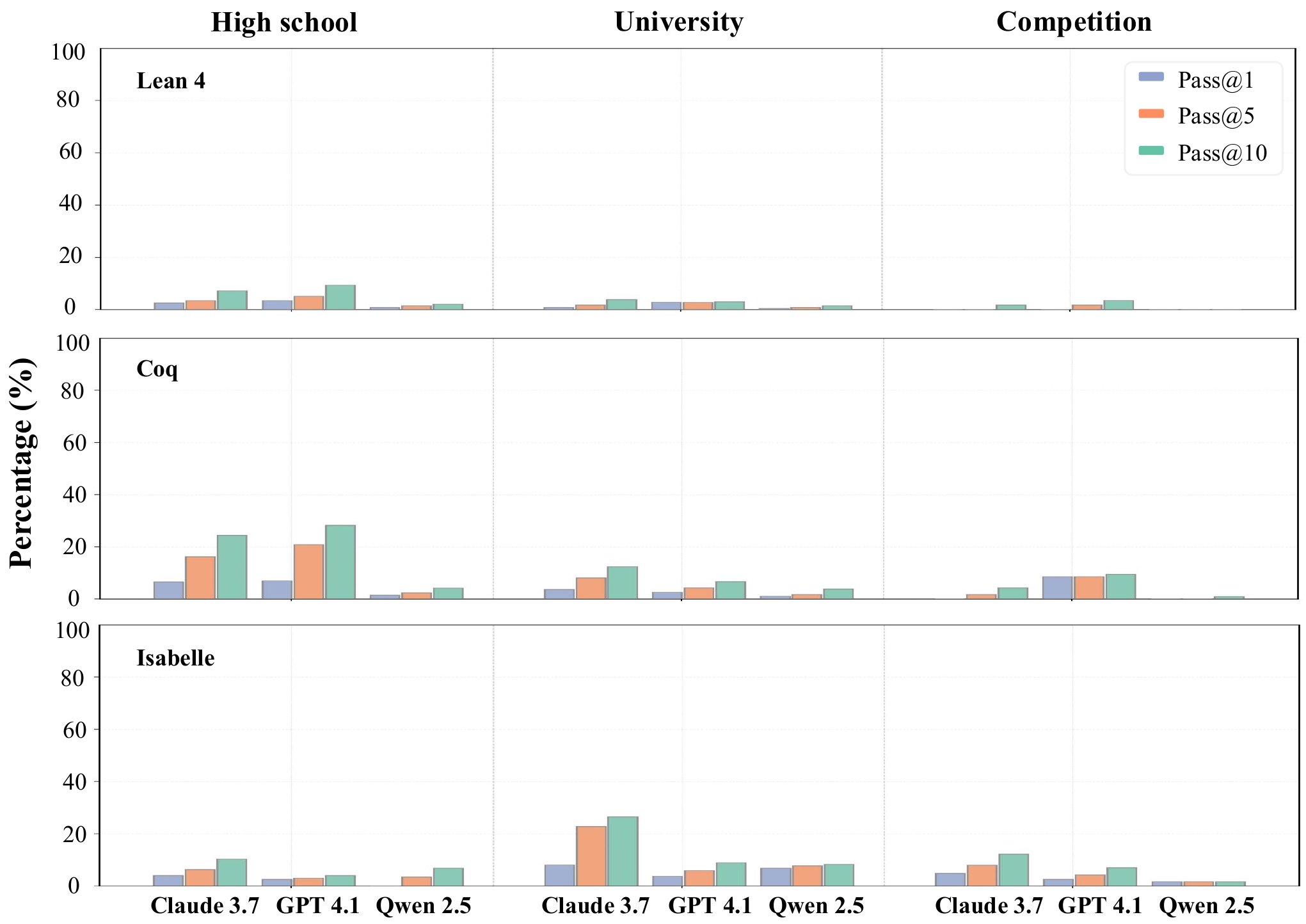}
    \caption{\small We present the performance of different MLLMs (Gemini-2.0-flash-thinking, OpenAI-GPT4.1, and Qwen2.5-VL-Instruct-70B) on multimodal automated theorem proving task across varying difficulty levels, evaluated using \textit{Pass@1, Pass@5, and Pass@10} metrics.}
    \label{fig:comp_language}
\end{figure} 

\section{Performance Comparison Across Formal Languages}
\label{appendix:comp}

Figure \ref{fig:comp_language} and the associated data illustrate the performance of Claude 3.7, GPT 4.1, and Qwen 2.5 on multimodal theorem proving tasks across Lean 4, Coq, and Isabelle, evaluated by Pass@1, Pass@5, and Pass@10 metrics across varying difficulty levels. The results consistently demonstrate that model performance significantly decreases with increasing task difficulty and improves with a greater number of allowed attempts, with Pass@10 achieving the highest pass rates. Among the evaluated models, GPT 4.1 generally exhibits the strongest performance across most formal languages and difficulty levels, particularly excelling in challenging scenarios. Claude 3.7 typically ranks as the second-best performer, while Qwen 2.5 consistently shows the lowest pass rates. Performance also varies by formal language, with models often achieving higher success rates in Coq compared to Lean 4 and Isabelle. The substantial difference between Pass@1 and Pass@10 highlights the models' ability to find correct proofs with multiple tries, although overall performance remains low on complex competition-level problems for all evaluated models.

\section{Limitations}
\label{limitation}
This paper evaluates the capabilities of various mainstream Multimodal Large Language Models (MLLMs) in multimodal automated theorem proving. We select three different formal languages—Lean 4, Coq, and Isabelle—for testing, and compare the performance (primarily using pass@10 as the metric) of general MLLMs including o1, Claude-3.7, Gemini-2.0, GPT-4.1, Qwen2.5-VL, and Llama3.2-Vision on problems of varying difficulty levels. Although this study provides an exploration into the application of MLLMs in the domain of formal proof, it also has several limitations. Firstly, the testing of MLLMs in this paper primarily involves one-shot generation of formal theorems and proofs, and does not explore multi-step or interactive proof generation capabilities. Secondly, the analysis of the model is primarily based on its final results, without delving into its internal mechanisms or the specific impact of different reasoning steps on performance. Future work could consider employing more comprehensive datasets, exploring richer evaluation scenarios such as multi-step and interactive proof generation, and conducting a more in-depth mechanistic analysis of the models' proof generation process.

\section{Prompts}
\label{prompt}
\begin{figure}[ht]
    \centering
    \includegraphics[width=\textwidth]{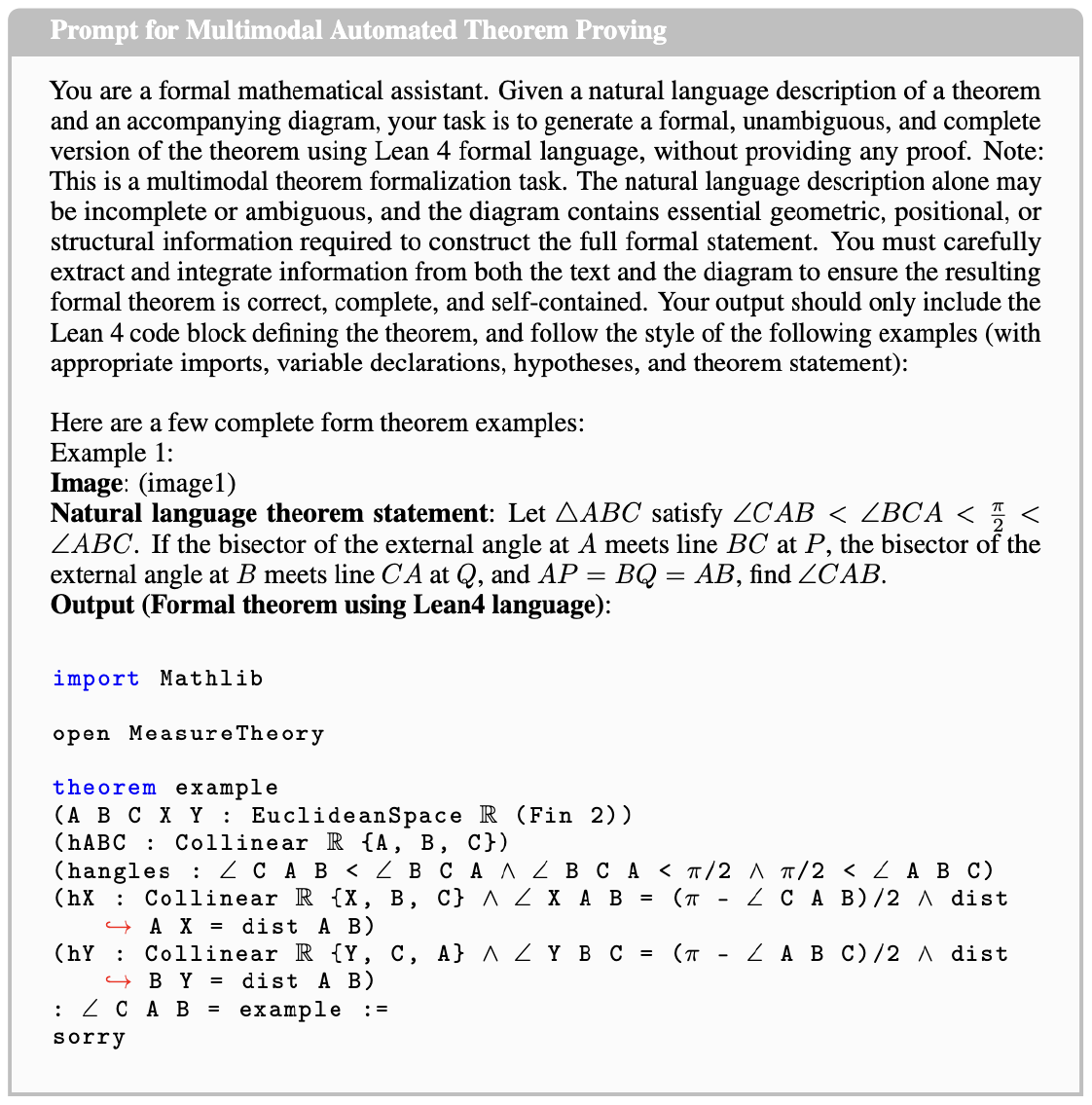}
    \caption{\small Prompt for multimodal automated theorem proving task (part1).}
    \label{fig:prompt1}
    \vspace{-0.3cm}
\end{figure}

\begin{figure}[ht]
    \centering
    \includegraphics[width=\textwidth]{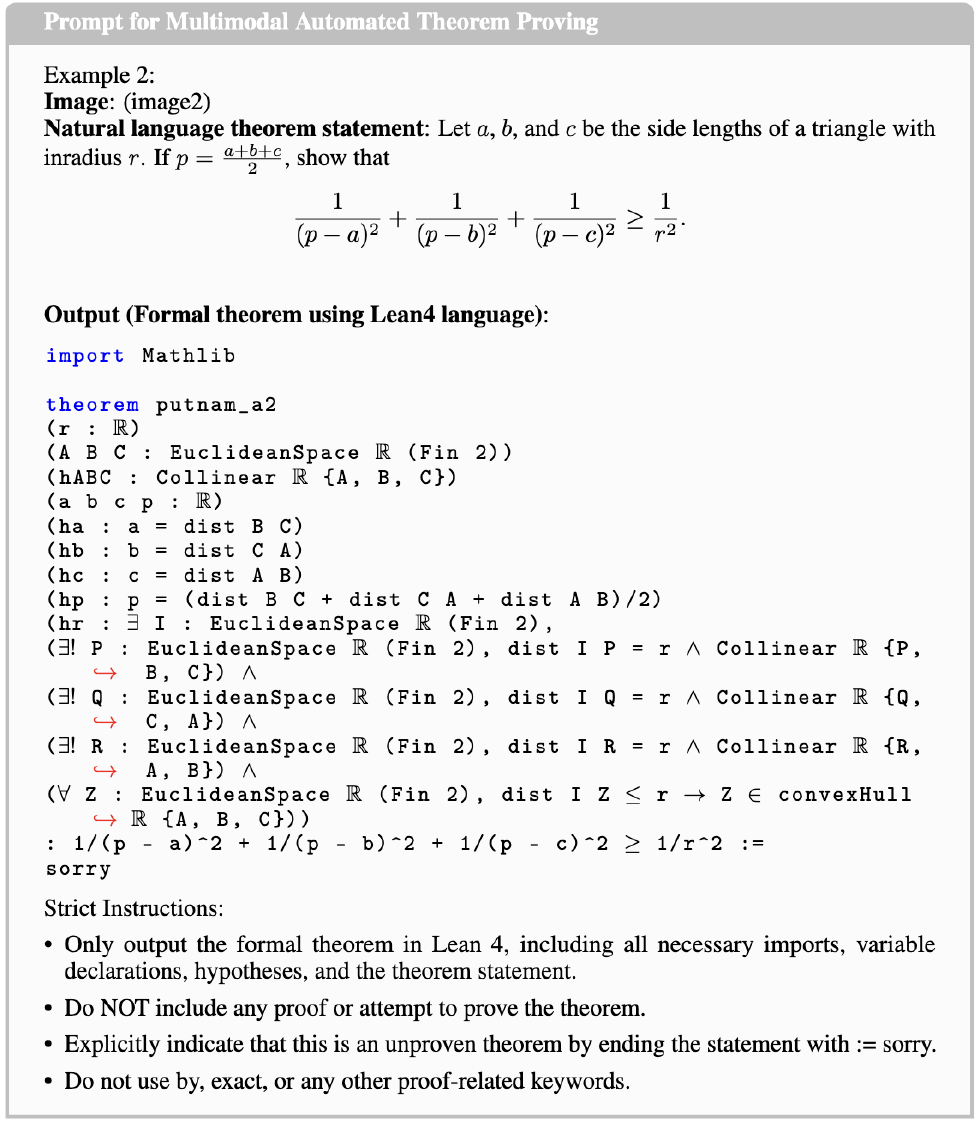}
    \caption{\small Prompt for multimodal automated theorem proving task (part2).}
    \label{fig:prompt1.1}
    \vspace{-0.3cm}
\end{figure}

\begin{figure}[ht]
    \centering
    \includegraphics[width=\textwidth]{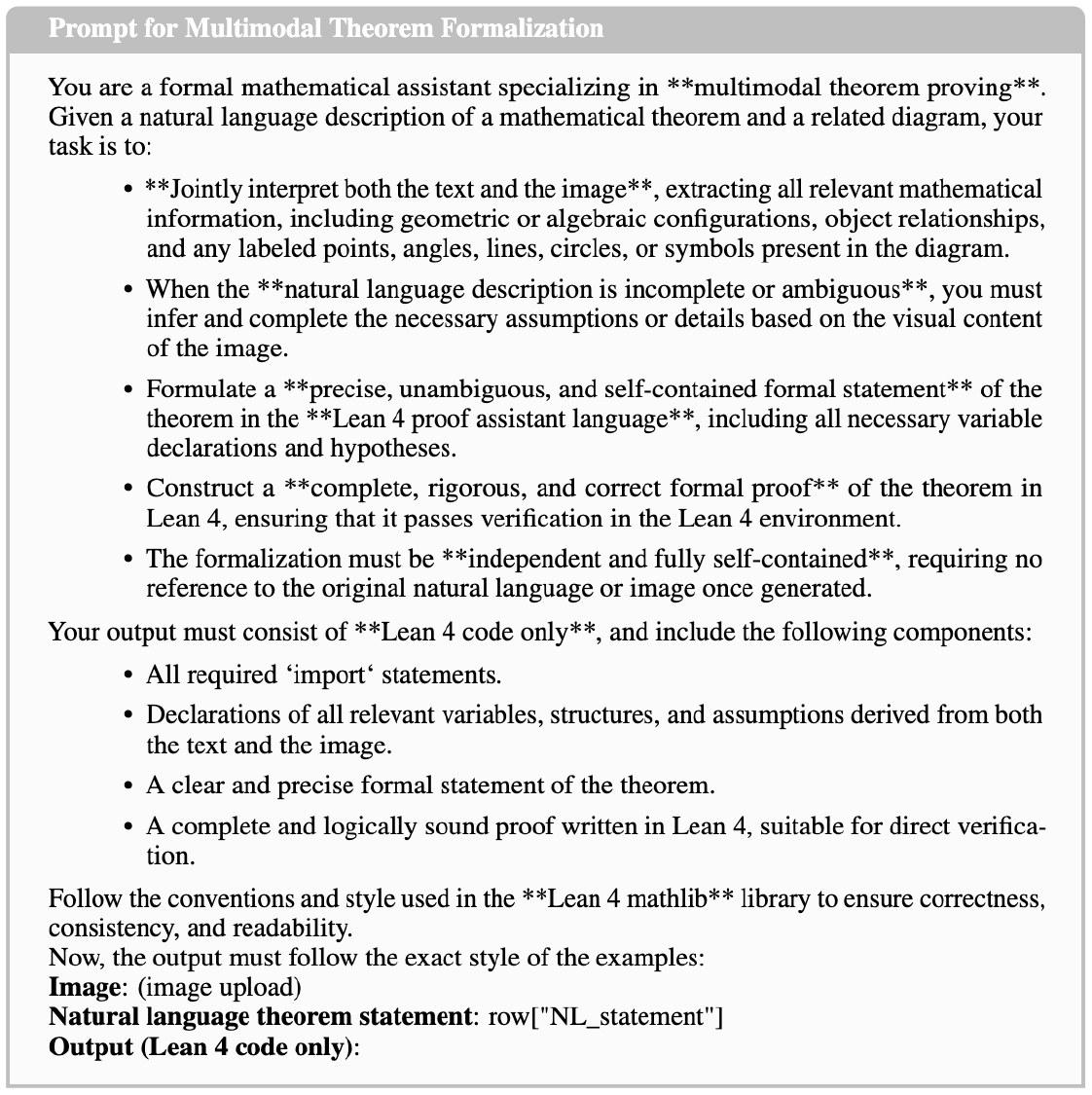}
    \caption{\small Prompt for multimodal theorem formalization task.}
    \label{fig:prompt2}
    \vspace{-0.3cm}
\end{figure}

Figure \ref{fig:prompt1} and \ref{fig:prompt1.1} outlines the prompt for multimodal automated theorem proving task, which aims to achieve end-to-end multimodal automated theorem proving similar to human provers, by directly generating a formalized theorem and its proof from multimodal informal input. Figure \ref{fig:prompt2} presents the prompt for multimodal theorem formalization task. The prover receives the multimodal question, and is required to formalize it into a precise theorem.


\newpage

\newpage

\newpage

\newpage
\newpage
\section*{NeurIPS Paper Checklist}

\begin{enumerate}

\item {\bf Claims}
    \item[] Question: Do the main claims made in the abstract and introduction accurately reflect the paper's contributions and scope?
    \item[] Answer: \answerYes{} 
    \item[] Justification: In this paper, we introduce  MATPBENCH, a new Multimodal, Multilingual, and Multi-level benchmark designed to evaluate MLLMs as automated theorem provers.
    \item[] Guidelines:
    \begin{itemize}
        \item The answer NA means that the abstract and introduction do not include the claims made in the paper.
        \item The abstract and/or introduction should clearly state the claims made, including the contributions made in the paper and important assumptions and limitations. A No or NA answer to this question will not be perceived well by the reviewers. 
        \item The claims made should match theoretical and experimental results, and reflect how much the results can be expected to generalize to other settings. 
        \item It is fine to include aspirational goals as motivation as long as it is clear that these goals are not attained by the paper. 
    \end{itemize}

\item {\bf Limitations}
    \item[] Question: Does the paper discuss the limitations of the work performed by the authors?
    \item[] Answer: \answerYes{} 
    \item[] Justification: Please refer to Section \ref{limitation}.
    \item[] Guidelines:
    \begin{itemize}
        \item The answer NA means that the paper has no limitation while the answer No means that the paper has limitations, but those are not discussed in the paper. 
        \item The authors are encouraged to create a separate "Limitations" section in their paper.
        \item The paper should point out any strong assumptions and how robust the results are to violations of these assumptions (e.g., independence assumptions, noiseless settings, model well-specification, asymptotic approximations only holding locally). The authors should reflect on how these assumptions might be violated in practice and what the implications would be.
        \item The authors should reflect on the scope of the claims made, e.g., if the approach was only tested on a few datasets or with a few runs. In general, empirical results often depend on implicit assumptions, which should be articulated.
        \item The authors should reflect on the factors that influence the performance of the approach. For example, a facial recognition algorithm may perform poorly when image resolution is low or images are taken in low lighting. Or a speech-to-text system might not be used reliably to provide closed captions for online lectures because it fails to handle technical jargon.
        \item The authors should discuss the computational efficiency of the proposed algorithms and how they scale with dataset size.
        \item If applicable, the authors should discuss possible limitations of their approach to address problems of privacy and fairness.
        \item While the authors might fear that complete honesty about limitations might be used by reviewers as grounds for rejection, a worse outcome might be that reviewers discover limitations that aren't acknowledged in the paper. The authors should use their best judgment and recognize that individual actions in favor of transparency play an important role in developing norms that preserve the integrity of the community. Reviewers will be specifically instructed to not penalize honesty concerning limitations.
    \end{itemize}

\item {\bf Theory assumptions and proofs}
    \item[] Question: For each theoretical result, does the paper provide the full set of assumptions and a complete (and correct) proof?
    \item[] Answer: \answerYes{} 
    \item[] Justification: Relevant assumptions and proofs are provided in the main paper.
    \item[] Guidelines:
    \begin{itemize}
        \item The answer NA means that the paper does not include theoretical results. 
        \item All the theorems, formulas, and proofs in the paper should be numbered and cross-referenced.
        \item All assumptions should be clearly stated or referenced in the statement of any theorems.
        \item The proofs can either appear in the main paper or the supplemental material, but if they appear in the supplemental material, the authors are encouraged to provide a short proof sketch to provide intuition. 
        \item Inversely, any informal proof provided in the core of the paper should be complemented by formal proofs provided in appendix or supplemental material.
        \item Theorems and Lemmas that the proof relies upon should be properly referenced. 
    \end{itemize}

    \item {\bf Experimental result reproducibility}
    \item[] Question: Does the paper fully disclose all the information needed to reproduce the main experimental results of the paper to the extent that it affects the main claims and/or conclusions of the paper (regardless of whether the code and data are provided or not)?
    \item[] Answer: \answerYes{} 
    \item[] Justification: See \url{https://github.com/Zhitao-He/MATPBench} and \url{https://matpbench.github.io/}.
    \item[] Guidelines:
    \begin{itemize}
        \item The answer NA means that the paper does not include experiments.
        \item If the paper includes experiments, a No answer to this question will not be perceived well by the reviewers: Making the paper reproducible is important, regardless of whether the code and data are provided or not.
        \item If the contribution is a dataset and/or model, the authors should describe the steps taken to make their results reproducible or verifiable. 
        \item Depending on the contribution, reproducibility can be accomplished in various ways. For example, if the contribution is a novel architecture, describing the architecture fully might suffice, or if the contribution is a specific model and empirical evaluation, it may be necessary to either make it possible for others to replicate the model with the same dataset, or provide access to the model. In general. releasing code and data is often one good way to accomplish this, but reproducibility can also be provided via detailed instructions for how to replicate the results, access to a hosted model (e.g., in the case of a large language model), releasing of a model checkpoint, or other means that are appropriate to the research performed.
        \item While NeurIPS does not require releasing code, the conference does require all submissions to provide some reasonable avenue for reproducibility, which may depend on the nature of the contribution. For example
        \begin{enumerate}
            \item If the contribution is primarily a new algorithm, the paper should make it clear how to reproduce that algorithm.
            \item If the contribution is primarily a new model architecture, the paper should describe the architecture clearly and fully.
            \item If the contribution is a new model (e.g., a large language model), then there should either be a way to access this model for reproducing the results or a way to reproduce the model (e.g., with an open-source dataset or instructions for how to construct the dataset).
            \item We recognize that reproducibility may be tricky in some cases, in which case authors are welcome to describe the particular way they provide for reproducibility. In the case of closed-source models, it may be that access to the model is limited in some way (e.g., to registered users), but it should be possible for other researchers to have some path to reproducing or verifying the results.
        \end{enumerate}
    \end{itemize}

\item {\bf Open access to data and code}
    \item[] Question: Does the paper provide open access to the data and code, with sufficient instructions to faithfully reproduce the main experimental results, as described in supplemental material?
    \item[] Answer: \answerYes{} 
    \item[] Justification: See \url{https://github.com/Zhitao-He/MATPBench}.
    \item[] Guidelines:
    \begin{itemize}
        \item The answer NA means that paper does not include experiments requiring code.
        \item Please see the NeurIPS code and data submission guidelines (\url{https://nips.cc/public/guides/CodeSubmissionPolicy}) for more details.
        \item While we encourage the release of code and data, we understand that this might not be possible, so “No” is an acceptable answer. Papers cannot be rejected simply for not including code, unless this is central to the contribution (e.g., for a new open-source benchmark).
        \item The instructions should contain the exact command and environment needed to run to reproduce the results. See the NeurIPS code and data submission guidelines (\url{https://nips.cc/public/guides/CodeSubmissionPolicy}) for more details.
        \item The authors should provide instructions on data access and preparation, including how to access the raw data, preprocessed data, intermediate data, and generated data, etc.
        \item The authors should provide scripts to reproduce all experimental results for the new proposed method and baselines. If only a subset of experiments are reproducible, they should state which ones are omitted from the script and why.
        \item At submission time, to preserve anonymity, the authors should release anonymized versions (if applicable).
        \item Providing as much information as possible in supplemental material (appended to the paper) is recommended, but including URLs to data and code is permitted.
    \end{itemize}

\item {\bf Experimental setting/details}
    \item[] Question: Does the paper specify all the training and test details (e.g., data splits, hyperparameters, how they were chosen, type of optimizer, etc.) necessary to understand the results?
    \item[] Answer: \answerYes{} 
    \item[] Justification: Please refer to Section \ref{experiment_setting}.
    \item[] Guidelines:
    \begin{itemize}
        \item The answer NA means that the paper does not include experiments.
        \item The experimental setting should be presented in the core of the paper to a level of detail that is necessary to appreciate the results and make sense of them.
        \item The full details can be provided either with the code, in appendix, or as supplemental material.
    \end{itemize}

\item {\bf Experiment statistical significance}
    \item[] Question: Does the paper report error bars suitably and correctly defined or other appropriate information about the statistical significance of the experiments?
    \item[] Answer: \answerYes{} 
    \item[] Justification: Please refer to Section \ref{results}.
    \item[] Guidelines:
    \begin{itemize}
        \item The answer NA means that the paper does not include experiments.
        \item The authors should answer "Yes" if the results are accompanied by error bars, confidence intervals, or statistical significance tests, at least for the experiments that support the main claims of the paper.
        \item The factors of variability that the error bars are capturing should be clearly stated (for example, train/test split, initialization, random drawing of some parameter, or overall run with given experimental conditions).
        \item The method for calculating the error bars should be explained (closed form formula, call to a library function, bootstrap, etc.)
        \item The assumptions made should be given (e.g., Normally distributed errors).
        \item It should be clear whether the error bar is the standard deviation or the standard error of the mean.
        \item It is OK to report 1-sigma error bars, but one should state it. The authors should preferably report a 2-sigma error bar than state that they have a 96\% CI, if the hypothesis of Normality of errors is not verified.
        \item For asymmetric distributions, the authors should be careful not to show in tables or figures symmetric error bars that would yield results that are out of range (e.g. negative error rates).
        \item If error bars are reported in tables or plots, The authors should explain in the text how they were calculated and reference the corresponding figures or tables in the text.
    \end{itemize}

\item {\bf Experiments compute resources}
    \item[] Question: For each experiment, does the paper provide sufficient information on the computer resources (type of compute workers, memory, time of execution) needed to reproduce the experiments?
    \item[] Answer: \answerYes{} 
    \item[] Justification: The experimental setup is detailed in Section \ref{experiment_setting}.
    \item[] Guidelines:
    \begin{itemize}
        \item The answer NA means that the paper does not include experiments.
        \item The paper should indicate the type of compute workers CPU or GPU, internal cluster, or cloud provider, including relevant memory and storage.
        \item The paper should provide the amount of compute required for each of the individual experimental runs as well as estimate the total compute. 
        \item The paper should disclose whether the full research project required more compute than the experiments reported in the paper (e.g., preliminary or failed experiments that didn't make it into the paper). 
    \end{itemize}
    
\item {\bf Code of ethics}
    \item[] Question: Does the research conducted in the paper conform, in every respect, with the NeurIPS Code of Ethics \url{https://neurips.cc/public/EthicsGuidelines}?
    \item[] Answer: \answerYes{} 
    \item[] Justification: We have reviewed the NeurIPS Code of Ethics and ensure our work aligns with its principles.
    \item[] Guidelines:
    \begin{itemize}
        \item The answer NA means that the authors have not reviewed the NeurIPS Code of Ethics.
        \item If the authors answer No, they should explain the special circumstances that require a deviation from the Code of Ethics.
        \item The authors should make sure to preserve anonymity (e.g., if there is a special consideration due to laws or regulations in their jurisdiction).
    \end{itemize}

\item {\bf Broader impacts}
    \item[] Question: Does the paper discuss both potential positive societal impacts and negative societal impacts of the work performed?
    \item[] Answer: \answerYes{} 
    \item[] Justification: Please refer to Section \ref{intro} and Section \ref{limitation}.
    \item[] Guidelines:
    \begin{itemize}
        \item The answer NA means that there is no societal impact of the work performed.
        \item If the authors answer NA or No, they should explain why their work has no societal impact or why the paper does not address societal impact.
        \item Examples of negative societal impacts include potential malicious or unintended uses (e.g., disinformation, generating fake profiles, surveillance), fairness considerations (e.g., deployment of technologies that could make decisions that unfairly impact specific groups), privacy considerations, and security considerations.
        \item The conference expects that many papers will be foundational research and not tied to particular applications, let alone deployments. However, if there is a direct path to any negative applications, the authors should point it out. For example, it is legitimate to point out that an improvement in the quality of generative models could be used to generate deepfakes for disinformation. On the other hand, it is not needed to point out that a generic algorithm for optimizing neural networks could enable people to train models that generate Deepfakes faster.
        \item The authors should consider possible harms that could arise when the technology is being used as intended and functioning correctly, harms that could arise when the technology is being used as intended but gives incorrect results, and harms following from (intentional or unintentional) misuse of the technology.
        \item If there are negative societal impacts, the authors could also discuss possible mitigation strategies (e.g., gated release of models, providing defenses in addition to attacks, mechanisms for monitoring misuse, mechanisms to monitor how a system learns from feedback over time, improving the efficiency and accessibility of ML).
    \end{itemize}
    
\item {\bf Safeguards}
    \item[] Question: Does the paper describe safeguards that have been put in place for responsible release of data or models that have a high risk for misuse (e.g., pretrained language models, image generators, or scraped datasets)?
    \item[] Answer: \answerNA{} 
    \item[] Justification: Our work does not involve the release of datasets or models with high risk for misuse.
    \item[] Guidelines:
    \begin{itemize}
        \item The answer NA means that the paper poses no such risks.
        \item Released models that have a high risk for misuse or dual-use should be released with necessary safeguards to allow for controlled use of the model, for example by requiring that users adhere to usage guidelines or restrictions to access the model or implementing safety filters. 
        \item Datasets that have been scraped from the Internet could pose safety risks. The authors should describe how they avoided releasing unsafe images.
        \item We recognize that providing effective safeguards is challenging, and many papers do not require this, but we encourage authors to take this into account and make a best faith effort.
    \end{itemize}

\item {\bf Licenses for existing assets}
    \item[] Question: Are the creators or original owners of assets (e.g., code, data, models), used in the paper, properly credited and are the license and terms of use explicitly mentioned and properly respected?
    \item[] Answer: \answerYes{} 
    \item[] Justification: Please refer to Section \ref{bench} and Section \ref{experiment_setting}.
    \item[] Guidelines:
    \begin{itemize}
        \item The answer NA means that the paper does not use existing assets.
        \item The authors should cite the original paper that produced the code package or dataset.
        \item The authors should state which version of the asset is used and, if possible, include a URL.
        \item The name of the license (e.g., CC-BY 4.0) should be included for each asset.
        \item For scraped data from a particular source (e.g., website), the copyright and terms of service of that source should be provided.
        \item If assets are released, the license, copyright information, and terms of use in the package should be provided. For popular datasets, \url{paperswithcode.com/datasets} has curated licenses for some datasets. Their licensing guide can help determine the license of a dataset.
        \item For existing datasets that are re-packaged, both the original license and the license of the derived asset (if it has changed) should be provided.
        \item If this information is not available online, the authors are encouraged to reach out to the asset's creators.
    \end{itemize}

\item {\bf New assets}
    \item[] Question: Are new assets introduced in the paper well documented and is the documentation provided alongside the assets?
    \item[] Answer: \answerYes{} 
    \item[] Justification: See \url{https://github.com/Zhitao-He/MATPBench}.
    \item[] Guidelines:
    \begin{itemize}
        \item The answer NA means that the paper does not release new assets.
        \item Researchers should communicate the details of the dataset/code/model as part of their submissions via structured templates. This includes details about training, license, limitations, etc. 
        \item The paper should discuss whether and how consent was obtained from people whose asset is used.
        \item At submission time, remember to anonymize your assets (if applicable). You can either create an anonymized URL or include an anonymized zip file.
    \end{itemize}

\item {\bf Crowdsourcing and research with human subjects}
    \item[] Question: For crowdsourcing experiments and research with human subjects, does the paper include the full text of instructions given to participants and screenshots, if applicable, as well as details about compensation (if any)? 
    \item[] Answer: \answerYes{} 
    \item[] Justification: Please refer to Section \ref{bench}.
    \item[] Guidelines:
    \begin{itemize}
        \item The answer NA means that the paper does not involve crowdsourcing nor research with human subjects.
        \item Including this information in the supplemental material is fine, but if the main contribution of the paper involves human subjects, then as much detail as possible should be included in the main paper. 
        \item According to the NeurIPS Code of Ethics, workers involved in data collection, curation, or other labor should be paid at least the minimum wage in the country of the data collector. 
    \end{itemize}

\item {\bf Institutional review board (IRB) approvals or equivalent for research with human subjects}
    \item[] Question: Does the paper describe potential risks incurred by study participants, whether such risks were disclosed to the subjects, and whether Institutional Review Board (IRB) approvals (or an equivalent approval/review based on the requirements of your country or institution) were obtained?
    \item[] Answer: \answerYes{} 
    \item[] Justification: Please refer to Section \ref{bench}.
    \item[] Guidelines:
    \begin{itemize}
        \item The answer NA means that the paper does not involve crowdsourcing nor research with human subjects.
        \item Depending on the country in which research is conducted, IRB approval (or equivalent) may be required for any human subjects research. If you obtained IRB approval, you should clearly state this in the paper. 
        \item We recognize that the procedures for this may vary significantly between institutions and locations, and we expect authors to adhere to the NeurIPS Code of Ethics and the guidelines for their institution. 
        \item For initial submissions, do not include any information that would break anonymity (if applicable), such as the institution conducting the review.
    \end{itemize}

\item {\bf Declaration of LLM usage}
    \item[] Question: Does the paper describe the usage of LLMs if it is an important, original, or non-standard component of the core methods in this research? Note that if the LLM is used only for writing, editing, or formatting purposes and does not impact the core methodology, scientific rigorousness, or originality of the research, declaration is not required.
    \item[] Answer: \answerNA{} 
    \item[] Justification: We only used LLM for writing, editing, or formatting.
    \item[] Guidelines:
    \begin{itemize}
        \item The answer NA means that the core method development in this research does not involve LLMs as any important, original, or non-standard components.
        \item Please refer to our LLM policy (\url{https://neurips.cc/Conferences/2025/LLM}) for what should or should not be described.
    \end{itemize}

\end{enumerate}

\end{document}